\documentclass[runningheads]{llncs}

\usepackage[T1]{fontenc}
\usepackage{graphicx}
\usepackage{comment}
\usepackage{amsmath,amssymb}
\usepackage{color}
\usepackage{url}
\usepackage[backref=page]{hyperref}
\usepackage{cite}
\usepackage[super]{nth}
\usepackage{booktabs}
\usepackage{makecell}
\usepackage{soul} 
\usepackage{multirow}
\usepackage{xspace}\usepackage{tabularx}
\newcolumntype{C}{>{\centering\arraybackslash}X}
\usepackage{colortbl}
\usepackage{pifont}
\usepackage[capitalise]{cleveref}
\usepackage{orcidlink}

\crefname{section}{Sec.}{Secs.}
\Crefname{section}{Section}{Sections}
\Crefname{table}{Table}{Tables}
\crefname{table}{Tab.}{Tabs.}
\Crefname{figure}{Figure}{Figures}
\crefname{figure}{Fig.}{Figs.}

\newcommand{\eg}{\textit{e.g.}\xspace}
\newcommand{\ie}{\textit{i.e.}\xspace}
\newcommand{\vs}{\textit{vs.}\xspace}

\newcommand*{\inparagraph}[1]{\medskip\noindent\textbf{#1}\hspace{0.5em}}

\definecolor{highlight}{RGB}{230, 230, 255}
\newcommand{\hcolor}{{\sethlcolor{highlight}\hl{Highlight}}}
\newcommand{\ours}{EMAT}
\newcommand{\cmark}{\ding{51}}
\DeclareRobustCommand{\nobreakdashc}{\nobreakdash} 
\begin{document}

\def\SubNumber{015}
\def\GCPRTrack{Main Track}

\title{Efficient Masked Attention Transformer for Few-Shot Classification and Segmentation}

\titlerunning{Efficient Masked Attention Transformer}

\author{Dustin Carrión-Ojeda\thanks{Corresponding author.}\inst{1,2}\orcidlink{0000-0001-5322-9130} \and
Stefan Roth\inst{1,2}\orcidlink{0000-0001-9002-9832} \and
Simone Schaub-Meyer\inst{1,2}\orcidlink{0000-0001-8644-1074}}

\authorrunning{D. Carrión-Ojeda et al.}

\institute{Department of Computer Science, Technical University of Darmstadt, Germany \and Hessian Center for AI (hessian.AI), Germany
\email{\{dustin.carrion,stefan.roth,simone.schaub\}@visinf.tu-darmstadt.de}\\
\url{https://visinf.github.io/emat}}

\maketitle

\begin{abstract}
Few-shot classification and segmentation (FS-CS) focuses on jointly performing multi-label classification and multi-class segmentation using few annotated examples. 
Although the current state of the art (SOTA) achieves high accuracy in both tasks, it struggles with small objects. 
To overcome this, we propose the \textbf{E}fficient \textbf{M}asked \textbf{A}ttention \textbf{T}ransformer (\ours), which improves classification and segmentation accuracy, especially for small objects. 
\ours{} introduces three modifications: a novel memory-efficient masked attention mechanism, a learnable downscaling strategy, and parameter-efficiency enhancements. 
\ours{} outperforms all FS-CS methods on the PASCAL-$5^i$ and COCO-$20^i$ datasets, using at least four times fewer trainable parameters. 
Moreover, as the current FS-CS evaluation setting discards available annotations, despite their costly collection, we introduce two novel evaluation settings that consider these annotations to better reflect practical scenarios.

\end{abstract}

\section{Introduction} 
Recently, data-intensive methods have been introduced for various deep learning applications \cite{Caron2021,Oquab2023,Kirillov2023,Radford2021,Gemini2023,Liu2023a,Chen2024}. 
These methods rely on large training datasets, making them impractical in fields where collecting extensive datasets is challenging or costly \cite{Fang2023,Zhu2023a,Fan2024}. 
Consequently, few-shot learning (FSL) methods have gained significant attention for their ability to learn from just a few examples and quickly adapt to new classes \cite{Tian2022,Wu2021,Ye2022,Aggarwal2023}. 
In computer vision, FSL has been mostly applied to image classification (FS-C)  \cite{Allen2019,Sun2019,Tian2020a,Herzog2024} and segmentation (FS-S) \cite{Dong2018,Moon2023,Zhu2024,Yang2023,Zhu2024b}. 

\begin{figure}[!t]
    \begin{tabularx}{\textwidth}{@{}CCCCCC@{}}
        GT Mask & CST\textsuperscript{*} \cite{Kang2023} & \ours & GT Mask & CST\textsuperscript{*} \cite{Kang2023} & \ours \\
        \includegraphics[width=0.16\textwidth]{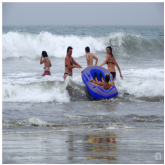} &
        \includegraphics[width=0.16\textwidth]{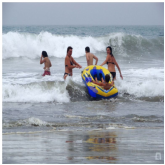} &
        \includegraphics[width=0.16\textwidth]{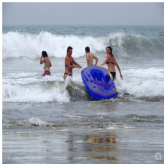} & 
        \includegraphics[width=0.16\textwidth]{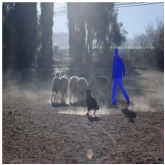} &
        \includegraphics[width=0.16\textwidth]{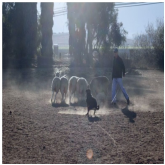} &
        \includegraphics[width=0.16\textwidth]{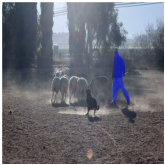}
    \end{tabularx}
    \vspace{-1.6em}
    \caption{\textbf{Qualitative comparison of small objects} (\ie, objects that occupy less than 15\,\% of the image) between the current SOTA FS-CS method (CST)~\cite{Kang2023} and our proposed \ours{}. 
    CST\textsuperscript{*} uses the same backbone as \ours{} (\ie, DINOv2~\cite{Oquab2023}). 
    By processing high-resolution correlation tokens, \ours{} preserves finer details, yielding more accurate segmentation masks.}
    \label{fig:teaser}
\end{figure}

FS-C and FS-S often co-occur in real-world applications, \eg, in agriculture, where crops must be segmented and classified by type or health status. 
Hence, recent works \cite{Kang2022,Kang2023} integrate multi-label classification and multi-class segmentation into a single few-shot classification and segmentation (FS-CS) task. 
While FS-CS addresses some limitations of FS-C (\eg, assuming the query image contains only one class) and FS-S (\eg, assuming the target class is always present in the query image), it also increases the task difficulty by simultaneously tackling classification and segmentation. 
Moreover, some applications, \eg, medical imaging, rely on precise small-object analysis \cite{Fan2024,Zhu2023a,Gong2021}. 
Thus, achieving high accuracy on small objects is a desired property for FS-CS methods. 
Yet, as shown in \cref{fig:teaser}, the current state-of-the-art (SOTA) FS-CS method  \cite{Kang2023} struggles with small objects, a limitation we address in this work.

To better align the evaluation of FS-CS models with practical scenarios, FS\nobreakdash-CS uses the \textit{N}-way \textit{K}-shot configuration, where the model learns \textit{N} classes from \textit{N}$\times$\textit{K} examples (\textit{K} per class). 
However, the current evaluation setting~\cite{Kang2022,Kang2023} discards available annotations, which is not ideal given the cost of data annotation. 
To address this, we introduce two new evaluation settings.

\subsubsection{Contributions.}
\emph{(1)} Building on the current SOTA FS-CS method \cite{Kang2023}, we propose an efficient masked attention transformer (\ours), which enhances classification and segmentation accuracy, particularly for small objects, while using approximately four times fewer trainable parameters.
\emph{(2)} Our \ours{} outperforms all FS-CS methods on the PASCAL-$5^i$ and COCO-$20^i$ datasets, supports the \textit{N}-way \textit{K}-shot configuration, and can generate empty segmentation masks when no target objects are present.
\emph{(3)} Finally, we introduce two new FS\nobreakdash-CS evaluation settings that better utilize available annotations during inference. 

\section{Related Work}
\subsubsection{Few-shot Classification (FS-C)} methods can be categorized into three groups based on what the model learns. 
\emph{Representation-based} approaches learn class-agnostic, discriminative embeddings \cite{Vinyals2016,Snell2017,Allen2019,Kang2021,Zhou2023,Hao2023,Ullah2022}. 
\emph{Optimization-based} approaches learn the optimal set of weights that allow the model to adapt to new classes in just a few optimization steps \cite{Finn2017,Sun2019,Antoniou2019,Raghu2020}. 
\emph{Transfer-based} approaches adapt large pre-trained \cite{Carrion2022,Singh2020,Tian2020a,Li2022,Ma2023} or foundation models \cite{Zhu2023b,Herzog2024,Silva-Rodriguez2024}.
A major limitation of most FS-C methods is the assumption of a single label per image \cite{Alfassy2019,Simon2022}, limiting them in multi-label settings. 

\subsubsection{Few-shot Segmentation (FS-S)} methods can also be categorized into three groups: 
\emph{prototype matching}, which aligns support embeddings with query features \cite{Dong2018,Wang2019,Tian2020c,Zhang2021,Liu2022,Wang2023}; \emph{dense correlation}, which constructs support–query correlation tensors \cite{Peng2023,Min2021,Xie2021a,Moon2023,Chen2024b,Xu2023}; and \emph{model-adaptation}, which fine-tunes large pre-trained models \cite{Zhou2024,Zhu2024,Wang2024,Liu2024,Zhang2024}. 
Despite the advancements in FS\nobreakdash-S, most methods have two main limitations: \emph{(1)} they target only the 1-way \textit{K}-shot configuration and \emph{(2)} they assume the query image contains the target class, preventing the models from predicting empty segmentation masks. 
Only a few recent works \cite{Tian2020b,Zhang2022} address the more general \textit{N}-way \textit{K}-shot configuration.

\subsubsection{Few-shot Classification and Segmentation (FS-CS)} focuses on jointly predicting the multi-label classification vector and multi-class segmentation mask without assuming support classes are present in the query image \cite{Kang2022}. 
The current SOTA FS-CS method, the classification-segmentation transformer (CST)~\cite{Kang2023}, uses a memory-intensive masked-attention mechanism that requires significant downsampling of the correlation features, reducing its accuracy on small objects. 
In this work, we enhance CST by proposing an efficient masked-attention formulation and adding further refinements, resulting in a more memory- and parameter-efficient method with improved accuracy, especially for small objects.

\section{Problem Definition} \label{sec:problem-definition}
This work focuses on the few-shot classification and segmentation (FS-CS) task \cite{Kang2022}, formulated as an \textit{N}-way \textit{K}-shot learning problem \cite{Vinyals2016}. 
We assume two disjoint class sets: $\mathcal{C}_\text{train}$ for training and $\mathcal{C}_\text{test}$ for testing. 
Accordingly, training tasks are sampled from $\mathcal{C}_\text{train}$, and testing tasks from $\mathcal{C}_\text{test}$. 
Each task consists of a support set $\mathcal{S}$ and a query image $\mathbf{I}_q$, where $\mathcal{S}$ contains \textit{N} classes $\mathcal{C}_\text{s}$ ($\mathcal{C}_\text{s} \subseteq \mathcal{C}_\text{train}$ or $\mathcal{C}_\text{s} \subseteq \mathcal{C}_\text{test}$), each represented by \textit{K} examples:
\begin{equation}
    \mathcal{S} = \left\{ \left\{(\mathbf{I}^i_j, \mathbf{M}^i_j, y^i_j) \mid y^i_j \in \mathcal{C}_\text{s} \right\}_j^K \right\}_i^N,
    \label{eq:original-setting}
\end{equation}
where $\mathbf{I}^i_j$, $\mathbf{M}^i_j$, and $y^i_j$ denote the support image, segmentation mask, and class label for the $j^\text{th}$ example of the $i^\text{th}$ class. 
Although $y^i_j = i \ \forall j$ in \cref{eq:original-setting}, we use this notation for compatibility with multi-label settings where $\mathbf{y}^i_j$ can vary. 

The goal of FS-CS is to learn from $\mathcal{S}$ such that, given  $\mathbf{I}_q$, the model can \emph{(i)}~identify which support classes are present (multi-label classification), and \emph{(ii)} segment those classes (multi-class segmentation). 
Moreover, FS-CS allows $\mathbf{I}_q$ to contain a subset of the support classes. 
Thus, when $N>1$, $\mathbf{I}_q$ can contain: \emph{(1)} none of the support classes ($\mathcal{C}_\text{q}=\varnothing$), \emph{(2)} a subset of them ($\mathcal{C}_\text{q} \subset \mathcal{C}_\text{s}$), or \emph{(3)}~all support classes ($\mathcal{C}_\text{q} = \mathcal{C}_\text{s}$). 
Note that case  \emph{(1)} is important in real-world applications where query images may not contain relevant classes, requiring models to predict empty segmentation masks when necessary.

\begin{figure}[!t]
    \centering
    \begin{tabularx}{\textwidth}{@{}ccCcCcCcC@{}}
        & \multicolumn{2}{c}{\scriptsize Available Data} & 
        \multicolumn{2}{c}{\scriptsize Original (Eq.\ \ref{eq:original-setting})} & 
        \multicolumn{2}{c}{\scriptsize Partially Aug. (Eq.\ \ref{eq:partially-augmented-setting})} & 
        \multicolumn{2}{c}{\scriptsize Fully Aug. (Eq.\ \ref{eq:fully-augmented-setting})} \\
        \cmidrule(lr){2-3} 
        \cmidrule(lr){4-5} 
        \cmidrule(lr){6-7}
        \cmidrule(lr){8-9}
        {\scriptsize $\mathbf{I}$} & 
        {\scriptsize $\mathbf{M}$} & 
        {\scriptsize $\mathbf{y}$} & 
        {\scriptsize $\mathbf{M}$} & 
        {\scriptsize $\mathbf{y}$} &  
        {\scriptsize $\mathbf{M}$} & 
        {\scriptsize $\mathbf{y}$} & 
        {\scriptsize $\mathbf{M}$} & 
        {\scriptsize $\mathbf{y}$} \\
        \includegraphics[width=0.16\textwidth]{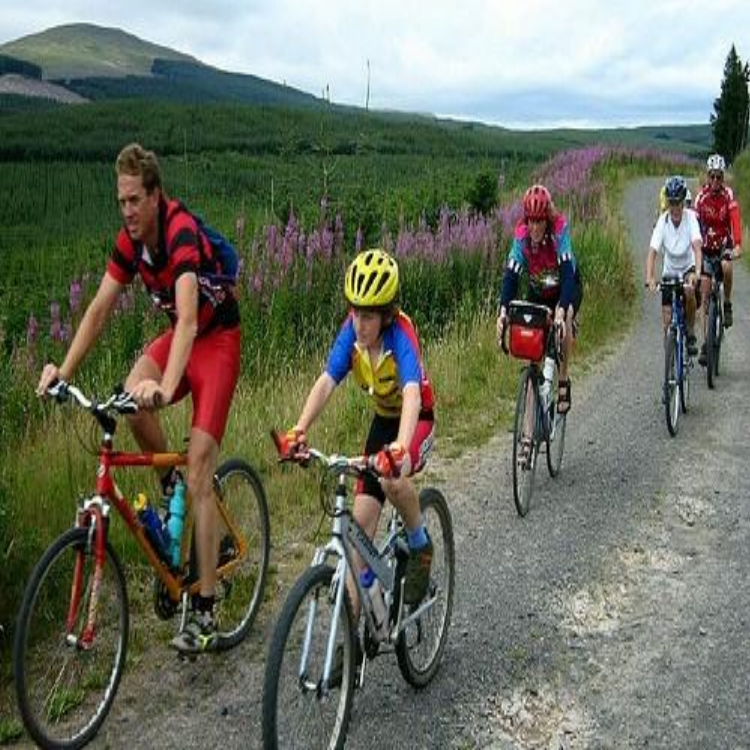} &
        \includegraphics[width=0.16\textwidth]{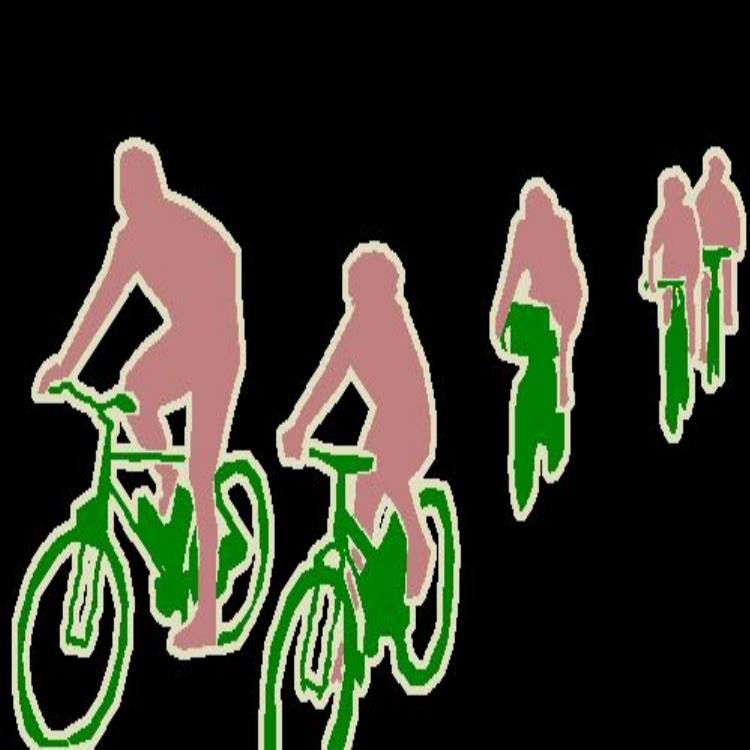} & {\vspace{-3.5em}\scriptsize\makecell{1 \\ 2}} &
        \includegraphics[width=0.16\textwidth]{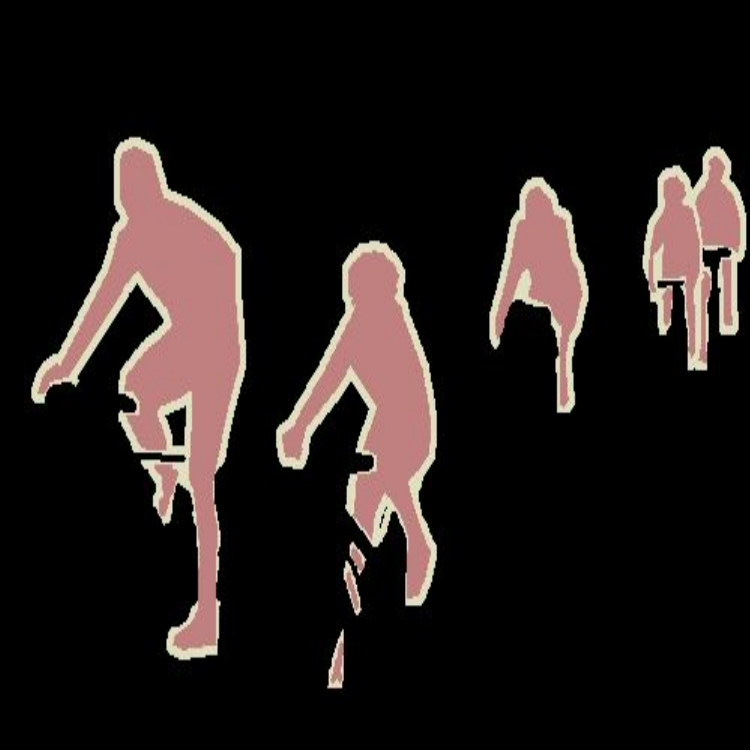} & {\vspace{-3.5em} \scriptsize 1} &
        \includegraphics[width=0.16\textwidth]{figures/eval-settings-mask1.pdf} & {\vspace{-3.5em}\scriptsize\makecell{1 \\ 2}} &
        \includegraphics[width=0.16\textwidth]{figures/eval-settings-mask1.pdf} & {\vspace{-3.5em}\scriptsize\makecell{1 \\ 2}} \\
        \includegraphics[width=0.16\textwidth]{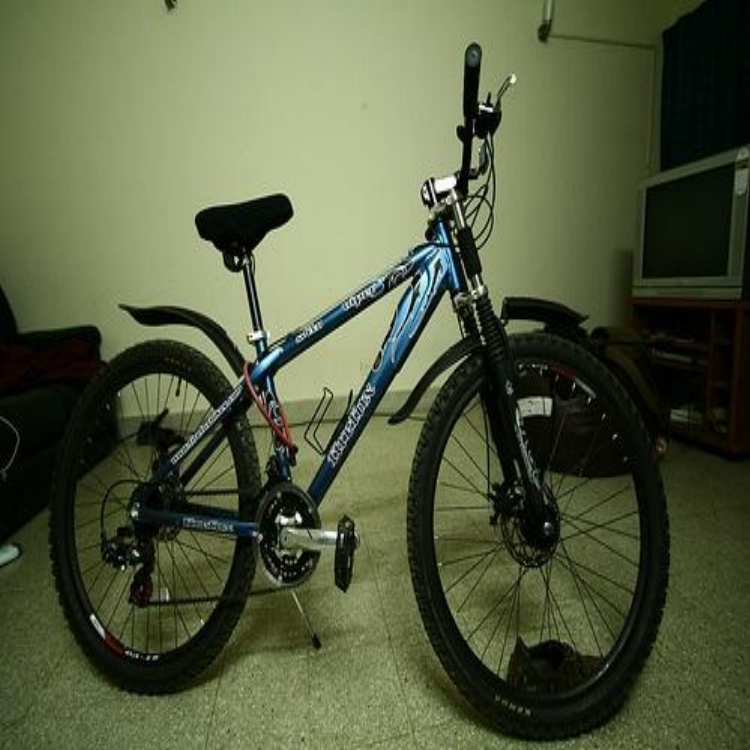} &
        \includegraphics[width=0.16\textwidth]{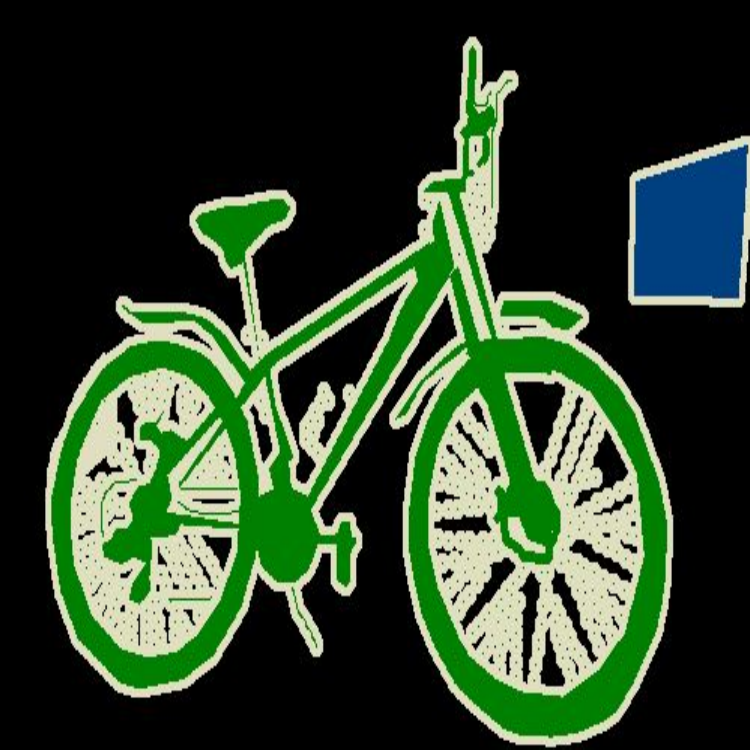} & {\vspace{-3.5em}\scriptsize\makecell{2 \\ 3}} &
        \includegraphics[width=0.16\textwidth]{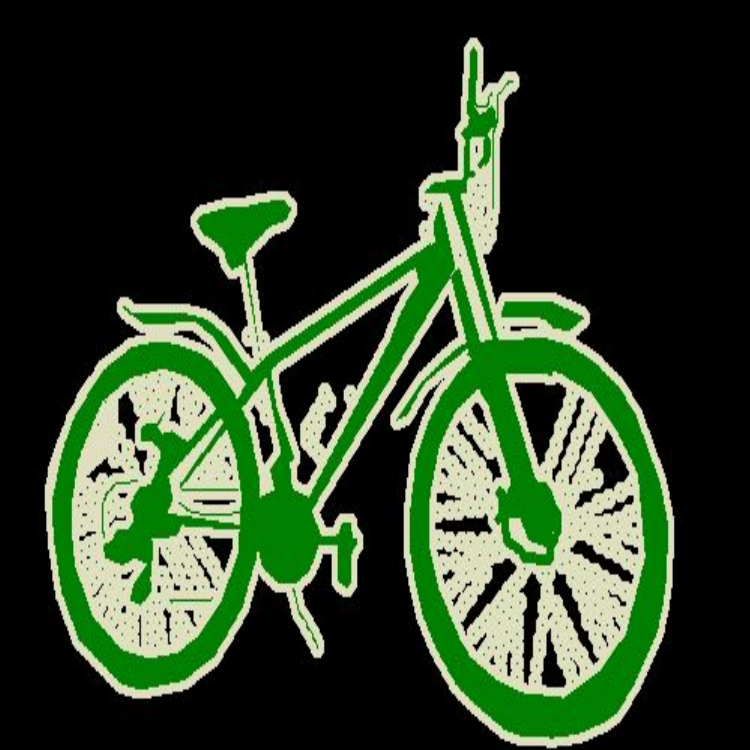} & {\vspace{-3.5em}\scriptsize 2} &
        \includegraphics[width=0.16\textwidth]{figures/eval-settings-mask2-removed.pdf} & {\vspace{-3.5em}\scriptsize 2} &
        \includegraphics[width=0.16\textwidth]{figures/eval-settings-mask2.pdf} & {\vspace{-3.5em}\scriptsize\makecell{2 \\ 3}} 
    \end{tabularx}
    \vspace{-1.4em}
    \caption{\textbf{Example of a 2-way 1-shot (base configuration) support set across different few-shot evaluation settings.} 
    $\mathbf{I}$, $\mathbf{M}$, and $\mathbf{y}$ represent the images, segmentation masks, and labels, respectively.} 
    \label{fig:fs-settings}
\end{figure}

The drawback of the current FS-CS setting is that each support image $\mathbf{I}^i_j$ is assumed to contain only one annotated class ($y^i_j$). 
If $\mathbf{I}^i_j$ includes multiple support classes ($\mathbf{y}^i_j \subseteq \mathcal{C}_\text{s}$), its label vector and segmentation mask need to be adjusted before constructing $\mathcal{S}$. 
This adjustment discards available annotations, as illustrated in \cref{fig:fs-settings}, where $\mathbf{I}_1^1$ contains both support classes (person, bike). 
However, since it is an example of the \nth{1} class (person), the annotations of the \nth{2} class (bike) are removed in the original setting. 

\subsection{Proposed Evaluation Settings} \label{sec:proposed-settings}
To better utilize available annotations and reflect more realistic evaluation scenarios, we introduce two novel FS-CS evaluation settings. 

\subsubsection{Partially Augmented Setting.} This setting keeps all annotations from the support classes: 
\begin{equation}
    \mathcal{S} = \left\{ \left\{(\mathbf{I}^i_j, \mathbf{M}^i_j, \mathbf{y}^i_j) \mid \mathbf{y}^i_j \subseteq \mathcal{C}_\text{s} \right\}_j^K \right\}_i^N.
    \label{eq:partially-augmented-setting}
\end{equation}
Note that when $N{=}1$, this setting is equivalent to \cref{eq:original-setting}. 
\cref{fig:fs-settings} shows an example for this setting, where $\mathbf{M}_1^1$ and $\mathbf{y}_1^1$ keep the annotations of the \nth{2} class (bike), even though the image is selected as an example of the \nth{1} class (person). 

\subsubsection{Fully Augmented Setting.} This setting keeps all available annotations for each support image, regardless of whether the corresponding classes are part of the support classes:
\begin{equation}
    \mathcal{S} = \left\{ \left\{(\mathbf{I}^i_j, \mathbf{M}^i_j, \mathbf{y}^i_j) \mid \mathbf{y}^i_j \subseteq \mathcal{C}_\text{train} \cup \mathcal{C}_\text{test} \right\}_j^K \right\}_i^N.
    \label{eq:fully-augmented-setting}
\end{equation}
For example, in \cref{fig:fs-settings}, $\mathbf{M}_1^1$ and $\mathbf{y}_1^1$ include annotations for both support classes (person, bike), while $\mathbf{M}_1^2$ and $\mathbf{y}_1^2$ are augmented with annotations of a non-support class (TV), which can belong to either $\mathcal{C}_\text{train}$ or $\mathcal{C}_\text{test}$. 
In this setting, the model is expected to classify and segment all support and augmented classes present in $\mathbf{I}_q$. 
Additionally, this setting aligns closely with the generalized few-shot setting (GFSL) \cite{Tian2022}, which also evaluates on base classes. 
However, unlike standard GFSL, which evaluates on all base classes seen during training, our setting restricts evaluation to only those classes present in the support set.

\section{Efficient Masked Attention Transformer} \label{sec:emat}
\begin{figure}[!t]
    \centering
    \includegraphics[width=\textwidth]{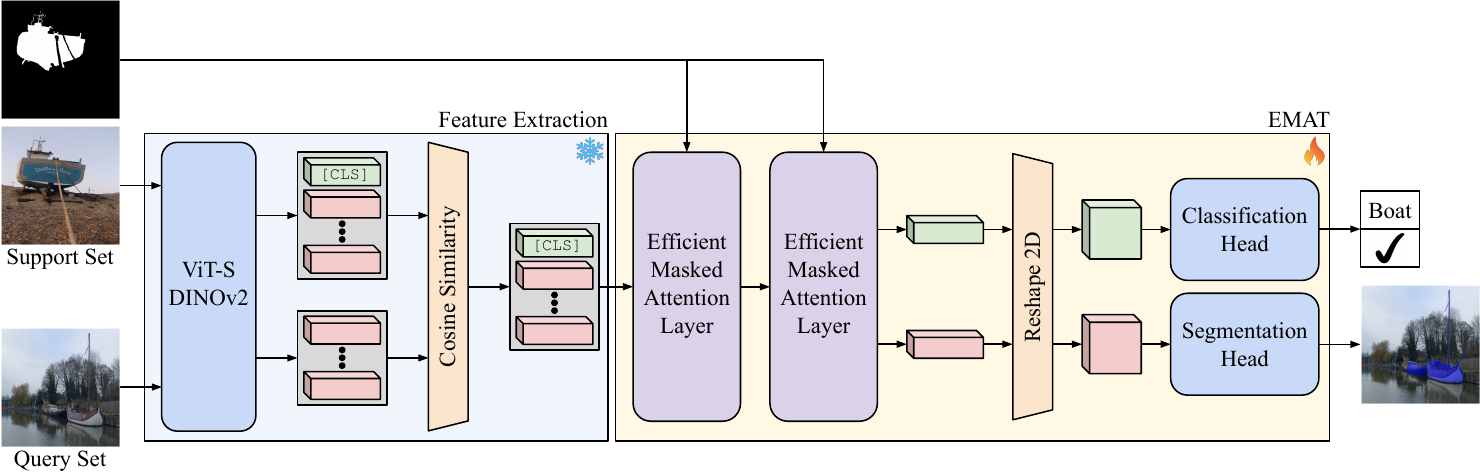}
    \vspace{-2.1em}
    \caption{\textbf{FS-CS pipeline used by our \ours{}.} 
    A frozen, pre-trained ViT \cite{Dosovitskiy2021} extracts image and class tokens from support and query images, which are correlated via cosine similarity. 
    The resulting correlation tokens are processed by a two-layer transformer equipped with our masked attention mechanism, learnable downscaling, and parameter-efficient design (see \cref{sec:memory-efficient-masked-attention,sec:learnable-downscaling,sec:parameter-efficiency}). 
    Task-specific heads then predict the multi-label classification vector and multi-class segmentation mask.} 
    \label{fig:emat}
\end{figure}

\cref{fig:emat} illustrates the pipeline used by our proposed efficient masked attention transformer (\ours{}), which builds upon the classification-segmentation transformer (CST) \cite{Kang2023}. 
Both methods share the same feature extraction process: support and query images $\mathbf{I}_j^i, \mathbf{I}_q \in \mathbb{R}^{H \times W \times 3}$ are processed by a frozen, pre-trained ViT \cite{Dosovitskiy2021} with patch size $p$, producing support and query image tokens $\mathbf{T}_{s_\text{i}}, \mathbf{T}_{q_\text{i}} \in \mathbb{R}^{h \times w \times d}$, and a support class token $\mathbf{T}_{s_\text{c}} \in \mathbb{R}^{1 \times d}$, where $h = H/p$, $w = W/p$, and $d$ is the token dimension of a single ViT head. 
The support tokens $\mathbf{T}_{s_\text{i}}$ are downsampled via bilinear interpolation and reshaped to $\mathbf{T}^f_{s_\text{i}} \in \mathbb{R}^{(h' \cdot w') \times d}$. 
Similarly, query image tokens $\mathbf{T}_{q_\text{i}}$ are reshaped to $\mathbf{T}^f_{q_\text{i}} \in \mathbb{R}^{(h \cdot w) \times d}$. 
Next, $\mathbf{T}^f_{s_\text{i}}$ and $\mathbf{T}_{s_\text{c}}$ are concatenated to form $\mathbf{T}^c_s$. 
Finally, cosine similarity between $\mathbf{T}^c_s$ and $\mathbf{T}^f_{q_\text{i}}$ is computed across all ViT layers $l$ and attention heads $g$, resulting in the correlation tokens $\mathbf{C} \in \mathbb{R}^{t_s \times t_q \times (l \cdot g)}$, where $t_s = h' \cdot w' + 1$ and $t_q=h \cdot w$. 

\ours{} differs from CST in the two-layer transformer (purple blocks in \cref{fig:emat}) that processes the correlation tokens $\mathbf{C}$ and feeds task-specific heads for multi-label classification and multi-class segmentation. 
\ours{} enhances this transformer with three key improvements: \emph{(1)} a novel memory-efficient masked attention formulation (see \cref{sec:memory-efficient-masked-attention}) that allows using higher-resolution correlation tokens, \emph{(2)} a learnable downscaling strategy (see \cref{sec:learnable-downscaling}) that avoids reliance on large pooling kernels, and \emph{(3)} additional modifications for improved parameter efficiency (see \cref{sec:parameter-efficiency}), which can help to reduce overfitting a small support set.

Following CST, \ours{} is trained using the 1-way 1-shot configuration. 
Since \ours{} uses task-specific heads, it is trained with two losses:
\begin{align}
    \mathcal{L}_{\text{clf}} &= -y\log \widehat{y},\label{eq:loss-clf}\\
    \mathcal{L}_{\text{seg}} &= -\frac{1}{HW}\sum_{i=1}^{H}{\sum_{j=1}^{W}{\mathbf{M}_{ij}\log \widehat{\mathbf{M}}_{ij}}},
    \label{eq:losses-clf-and-seg}
\end{align}
where $y\in \{0,1\}$ and $\mathbf{M}_{ij} \in \{0,1\}$ are the ground-truth classification and segmentation labels, and $\widehat{y}$, $\widehat{\mathbf{M}}_{ij}$ are the corresponding predictions. 
The final loss function jointly optimizes both losses using a balancing hyperparameter $\lambda$:
\begin{equation}
    \mathcal{L} = \lambda\mathcal{L}_{\text{clf}} + \mathcal{L}_{\text{seg}} .
    \label{eq:full-loss}
\end{equation}
Inference on \textit{N}-way \textit{K}-shot task is performed as in CST \cite{Kang2023}, by treating each class as an independent 1-way \textit{K}-shot task: class-wise logits and segmentation masks are averaged over the \textit{K} examples, producing \textit{N} predictions. 
Logits above a threshold $\delta = 0.5$ form the multi-label vector, and each pixel $\widehat{\mathbf{M}}_{ij}$ is assigned to the class with the highest score, or to background if all scores fall below $\delta$, thereby allowing empty masks.

\subsection{Memory-Efficient Masked Attention} \label{sec:memory-efficient-masked-attention}
The SOTA FS-CS method, CST \cite{Kang2023}, uses a masked attention mechanism based on self-attention \cite{Vaswani2017,Dosovitskiy2021} to process the correlation tokens $\mathbf{C} \in \mathbb{R}^{t_s \times t_q \times (l \cdot g)}$. 
Due to the high memory cost of self-attention, CST significantly downsamples the support tokens $\mathbf{T}_{s_\text{i}}$ when computing $\mathbf{C}$, sacrificing fine-grained spatial details (see \cref{fig:mask-comparison}). 
To address this, we propose a novel memory-efficient masked attention formulation that allows \ours{} to use high-resolution correlation tokens. 

Given $\mathbf{C}$, let $\mathbf{Q}^d \in \mathbb{R}^{t_d \times t_q \times e}$, $\mathbf{K}, \mathbf{V} \in \mathbb{R}^{t_s \times t_q \times e}$ denote the query, key, and value matrices, where $e$ is the embedding size. 
The dimensions of $\mathbf{Q}^d$ differ from those of $\mathbf{K}$ and $\mathbf{V}$ because the query matrix is progressively downscaled (see \cref{sec:learnable-downscaling}). 
Additionally, as shown in \cref{fig:emat}, the segmentation mask $\mathbf{M} \in \mathbb{R}^{H \times W}$ enters directly into the attention mechanism without being processed by the feature extractor. 
Thus, it is resized, flattened, and then append a trailing ``1'' to obtain $\mathbf{M}^f \in \mathbb{R}^{h' \cdot w' + 1}$. This appended ``1'' ensures that the support class token is never masked in \cref{eq:masked-attention,eq:efficient-masked-attention}.

For one attention head, CST computes: 
\begin{equation}
    \mathbf{O}_{ijk} = \sum_p \left[\mathrm{softmax} \left( \left( \mathbf{Q}_{ijk}^d \cdot \mathbf{K}_{:jk} \right) \odot \mathbf{M}^f_{:} \right)\right]_p \odot \mathbf{V}_{pjk},
    \label{eq:masked-attention}
\end{equation}
where $i \in \{1,\dots,t_d\}$, $j \in \{1,\dots, t_q \}$, $k \in \{1,\dots, e \}$, and $p \in \{1,\dots,t_s\}$. 
Here $t_d = h'' \cdot w'' + 1$, $t_q = h \cdot w$, and $t_s = h' \cdot w' + 1$, the ``$+1$'' accounts for the support class token, and $(\cdot)'$ denotes a downscaled value. 
The operator $\odot$ represents element-wise multiplication. 

In CST, the support dimension $t_s$ is 145 ($h' = w' = 12$) in the first attention layer and 10 ($h' = w' = 3$) in the second. Consequently, in the second layer most values of $\mathbf{M}^f$ are zero (see \cref{fig:mask-comparison}). As a result, \cref{eq:masked-attention} zeros out most attention values, but they are still processed in all intermediate computations, leading to a memory-inefficient formulation. To overcome this, we introduce a memory-efficient reformulation that excludes the masked-out entries:
\begin{equation}
    \mathbf{O}_{ijk} = \sum_{p^\oslash} \left[ \mathrm{softmax} \left( \mathbf{Q}_{ijk}^d \cdot \left( \mathbf{K}_{:jk} \oslash \mathbf{M}^f_{:}  \right) \right) \right]_{p^\oslash} \odot \left( \mathbf{V}_{:jk} \oslash \mathbf{M}^f_{:} \right)_{p^\oslash},
    \label{eq:efficient-masked-attention}
\end{equation}
where $\oslash$ is our element-wise masking operator:
\begin{equation}
    (\mathbf{Z}_{pjk} \oslash \mathbf{M}^f_{p}) = \begin{cases}
    \mathbf{Z}_{pjk} & \text{if} \; \mathbf{M}^f_{p} = 1, \\
    \varnothing & \text{otherwise},
    \end{cases}
    \qquad \forall p \in \{1,\dots,t_s\} ,
    \label{eq:element-wise-masking}
\end{equation}
with $\mathbf{Z}\in \mathbb{R}^{t_s \times t_q \times e}$ and $\varnothing$ indicating that the corresponding entry is excluded. 
This exclusion of elements results in the reduced set of indices $p^\oslash \subseteq p$ used in \cref{eq:efficient-masked-attention}, where $p^\oslash=p$ only if  $\mathbf{M}^f$ contains no zeros. 
Excluding masked-out tokens reduces memory usage allowing \ours{} to increase the support dimension $t_s$ to 401 ($h' = w' = 20$) in the first attention layer ($\approx 2.7$ times more than CST) and 101 ($h' = w' = 10$) in the second ($\approx 11$ times more than CST). 
Note that the index set $p^\oslash$ varies across images in a batch; thus, \cref{eq:efficient-masked-attention} is computed sequentially for each image. 
However, batch processing is still used both before and after this step because the input and output tensors ($\mathbf{C}$ and $\mathbf{O}$) have the same dimensions for every image. 
By computing attention only over unmasked entries, \ours{} still achieves a runtime comparable to CST.

\subsection{Learnable Downscaling} \label{sec:learnable-downscaling}
As mentioned in \cref{sec:memory-efficient-masked-attention}, the query matrix $\mathbf{Q} \in \mathbb{R}^{t_s \times t_q \times e}$ is progressively downscaled. 
This downscaling occurs before computing the masked-attention in each layer and it keeps $t_q$ and $e$ fixed, while shrinking the support spatial dimensions $h'$ and $w'$, which reduces the support dimension ($t_s = h'\cdot w'+1$).
Unlike CST, which uses only average pooling, \ours{} introduces a lightweight, learnable strategy that combines small convolutions with pooling. 
This hybrid design removes the need for the large pooling kernels that would otherwise be required to handle the higher-resolution correlation tokens used by \ours{}.

In the first attention layer, \ours{} splits $\mathbf{Q}$ into support image tokens $\mathbf{Q}_i$ and a single class token $\mathbf{Q}_c$. 
After reshaping $\mathbf{Q}_i$ to its $h'\times w'$ support spatial layout, a 3D convolution followed by another reshape produces $\mathbf{Q}^r_i \in \mathbb{R}^{(h''\cdot w'') \times t_q \times e}$. 
Simultaneously, a 2D convolution transforms $\mathbf{Q}_c$ into $\mathbf{Q}^r_c \in \mathbb{R}^{1 \times t_q \times e}$. 
These outputs are concatenated to form the downscaled query matrix $\mathbf{Q}^d \in \mathbb{R}^{t_d \times t_q \times e}$ used in \cref{eq:efficient-masked-attention}, where $t_d = h''\cdot w'' +1$.

The second attention layer repeats the process, but before the concatenation $\mathbf{Q}^r_i$ is collapsed to a single spatial token by a 3D average pool, producing $\mathbf{Q}^p_i \in \mathbb{R}^{1 \times t_q \times e}$. 
Concatenating $\mathbf{Q}^p_i$ with $\mathbf{Q}^r_c$ gives the downscaled query matrix $\mathbf{Q}^d \in \mathbb{R}^{2 \times t_q \times e}$ used during the attention computation.

\subsection{Modifications for Parameter Efficiency} \label{sec:parameter-efficiency}
Few-shot models with too many parameters risk overfitting the small support set, thereby reducing their ability to adapt to new classes. Therefore, \ours{} reduces the number of channels across all operations: its two attention layers use 64 and 32 channels, versus 32 and 128 in CST, and its two task-specific heads use 32 and 16 channels, versus 128 and 64 in CST. These channel reductions significantly decrease the number of trainable parameters in \ours{}.

\section{Experiments}

\subsubsection{Datasets.} We evaluated our \ours{} on the widely used PASCAL-$5^i$ \cite{Shaban2017} and COCO-$20^i$ \cite{Nguyen2019} datasets. 
Although they were designed for few-shot segmentation, both can also be used for few-shot classification and segmentation \cite{Kang2022}. 
PASCAL\nobreakdash-$5^i$ comprises 20 classes and COCO-$20^i$ 80 classes, each partitioned into four non-overlapping folds. 

\subsubsection{Implementation Details.} \ours{} uses a frozen ViT-S encoder \cite{Dosovitskiy2021} pre-trained with DINOv2 \cite{Oquab2023}. 
The two-layer transformer uses our memory-efficient masked attention with 8 heads. 
We train for 80 epochs with a batch size of 9 using the Adam optimizer \cite{kingma2015} with learning rate $10^{-3}$. 
Following \cite{Kang2023}, we use 1-way 1-shot tasks with the original setting (see Eq.\ \ref{eq:original-setting}) and set the loss weight $\lambda$ in \cref{eq:full-loss} to 0.1. 
Moreover, we re-train CST \cite{Kang2023} with the same DINOv2 backbone used by \ours{} and denote it as CST\textsuperscript{*}. 
All training was conducted on three NVIDIA RTX A6000 GPUs, with evaluation performed on a single GPU.

\subsection{Comparison to SOTA FS-CS} \label{sec:comparison-sota-fscs}
To evaluate the effectiveness of our \ours, we compare it with CST \cite{Kang2023} and other state-of-the-art few-shot classification and segmentation (FS-CS) methods. 
\cref{tab:comparison-fscs} shows the mean classification accuracy (Acc.) and mean Intersection over Union (mIoU) over the four folds of PASCAL-$5^i$ \cite{Shaban2017} and COCO-$20^i$ \cite{Nguyen2019}, for 2-way 1-shot tasks across all evaluation settings (see \cref{sec:problem-definition}). 
Although DINOv2 pre-training \cite{Oquab2023} already significantly improves CST\textsuperscript{*} over its original version, \ours{} consistently outperforms all methods across all settings. 
These results validate the benefit of processing higher-resolution correlation tokens enabled by our memory-efficient masked attention (see \cref{sec:memory-efficient-masked-attention}). 
Moreover, \ours{} requires at least four times fewer parameters than CST, making it the most parameter-efficient method among SOTA FS-CS models. 
The supplementary material provides per-fold results for \cref{tab:comparison-fscs} and additional results on $\{1,\dots,5\}$-way 5-shot tasks, demonstrating the scalability of our method.

The results in \cref{tab:comparison-fscs} also show that our partially augmented setting slightly improves accuracy and mIoU for most methods, confirming the benefit of better exploiting the available annotations. 
However, the improvement is marginal, likely because only 242 and 106 out of 4000 tasks are augmented for PASCAL\nobreakdash-$5^i$ and COCO-$20^i$, respectively. 
In contrast, our fully augmented setting lowers accuracy and mIoU for every method, although less significantly for mIoU. 
As discussed in \cref{sec:proposed-settings}, this setting augments not only the support examples but also includes every class present in the support images, making the tasks harder. 
This setting augments 1243 and 1515 out of the 4000 tasks for PASCAL-$5^i$ and COCO-$20^i$, respectively. 
The augmentations in both of our proposed settings highlight that the original evaluation setting fails to use available annotations.

\begin{table}[!t]
    \centering
    \caption{\textbf{Comparison of FS-CS methods} on PASCAL-$5^i$ and COCO-$20^i$ across all evaluation settings: original, partially augmented, and fully augmented, using 2\nobreakdashc-way 1-shot tasks (base configuration). 
    CST\textsuperscript{*} and \ours{} were trained and evaluated, while other methods were only evaluated using the checkpoints from \cite{Kang2022}. 
    CST\textsuperscript{*} uses the same backbone as \ours{} (\ie, DINOv2 \cite{Oquab2023}). 
    All values, except the number of trainable parameters (in millions), are percentages (higher is better). 
    \hcolor{} indicates our proposed method. 
    \textbf{Bold} and \underline{underlined} values indicate the best and second best results.}
    \label{tab:comparison-fscs}
    \vspace{-0.5em}
    \begin{tabularx}{\textwidth}{l@{\hspace{0.5em}}lrCCCCCC}
        \toprule
        \multirow{2.5}{*}{\textbf{Dataset}} &
        \multirow{2.5}{*}{\textbf{Method}} &
        \multirow{2.5}{*}{{\makecell{\textbf{Train.} \\ \textbf{Params.}}}} &
        \multicolumn{2}{c}{\textbf{Original}} &
        \multicolumn{2}{c}{{\makecell{\textbf{Partially} \\ \textbf{Augmented}}}} &
        \multicolumn{2}{c}{{\makecell{\textbf{Fully} \\ \textbf{Augmented}}}} \\
        \cmidrule(lr){4-5} 
        \cmidrule(lr){6-7} 
        \cmidrule(lr){8-9} 
        & & &
        \textbf{Acc.} & \textbf{mIoU} &
        \textbf{Acc.} & \textbf{mIoU} &
        \textbf{Acc.} & \textbf{mIoU} \\ 
        \midrule
        \multirow{7}{*}{PASCAL-$5^i$} & PANet \cite{Wang2019} & 23.51 & 56.53 & 37.20 & 56.93 & 37.49 & 55.75 & 37.25 \\
        & PFENet \cite{Tian2020c} & 31.96 & 39.35 & 35.57 & 39.48 & 35.61 & 36.88 & 35.08 \\
        & HSNet \cite{Min2021} & 2.57 & 67.27 & 44.85 & 67.75 & 44.72 & 65.92 & 44.40 \\
        & ASNet \cite{Kang2022} & 1.32 & 68.30 & 47.87 & 68.62 & 47.78 & 66.40 & 47.58 \\
        & CST \cite{Kang2023} & \underline{0.37} & 70.37 & 53.78 & 70.60 & 53.81 & 68.45 & 53.76 \\
        & CST\textsuperscript{*} & \underline{0.37} & \underline{80.58} & \underline{63.28} & \underline{80.60} & \underline{63.23} & \underline{78.57} & \underline{63.08} \\
        \rowcolor{highlight}
        \cellcolor{white}& \ours & \bfseries 0.09 & \bfseries 82.70 & \bfseries 63.38 & \textbf{82.92} & \bfseries 63.32 & \bfseries 81.23 & \bfseries 63.24 \\
        \midrule
        \multirow{7}{*}{COCO-$20^i$} & PANet \cite{Wang2019} & 23.51 & 51.30 & 23.64 & 51.32 & 23.78 & 45.07 & 23.17 \\
        & PFENet \cite{Tian2020c} & 31.96 & 36.45 & 23.37 & 36.50 & 23.39 & 29.33 & 21.61 \\
        & HSNet \cite{Min2021} & 2.57 & 62.43 & 30.58 & 62.40 & 30.66 & 55.15 & 29.44 \\
        & ASNet \cite{Kang2022} & 1.32 & 63.05 & 31.62 & 63.03 & 31.64 & 55.47 & 30.47 \\
        & CST \cite{Kang2023} & \underline{0.37} & 64.02 & 36.23 & 64.10 & 36.20 & 56.30 & 35.60 \\
        & CST\textsuperscript{*} & \underline{0.37} & \underline{78.70} & \underline{51.47} & \underline{78.87} & \underline{51.53} & \underline{71.18} & \underline{50.76} \\
        \rowcolor{highlight}
        \cellcolor{white}& \ours & \bfseries 0.09 & \bfseries 80.07 & \bfseries 52.81 & \textbf{80.25} & \bfseries 52.82 & \bfseries 73.00 & \bfseries 51.99 \\ 
        \bottomrule
    \end{tabularx}
\end{table}

\subsection{Qualitative Results}
\begin{figure}[!t]
    \centering
    \begin{tabularx}{\textwidth}{@{}CCCCCC@{}}
        Support ($\mathbf{I}_1^1$) & 
        Support ($\mathbf{I}_1^2$) & 
        Query ($\mathbf{I}_q$) &
        GT Mask &
        CST\textsuperscript{*} \cite{Kang2023} &
        \ours{} \\
        \includegraphics[width=0.16\textwidth]{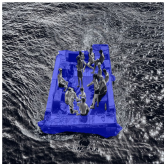} &
        \includegraphics[width=0.16\textwidth]{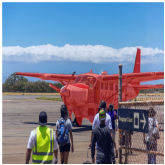} &
        \includegraphics[width=0.16\textwidth]{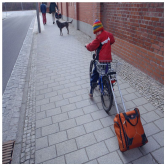} &
        \includegraphics[width=0.16\textwidth]{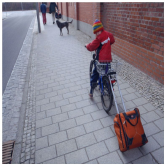} &
        \includegraphics[width=0.16\textwidth]{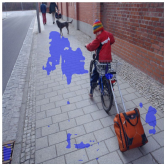} &
        \includegraphics[width=0.16\textwidth]{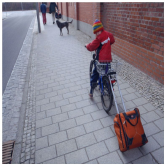} \\
        \includegraphics[width=0.16\textwidth]{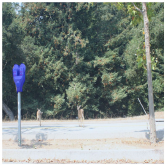} &
        \includegraphics[width=0.16\textwidth]{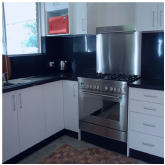} &
        \includegraphics[width=0.16\textwidth]{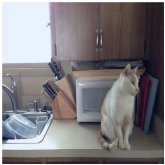} &
        \includegraphics[width=0.16\textwidth]{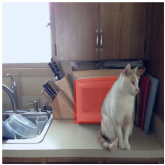} &
        \includegraphics[width=0.16\textwidth]{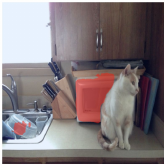} &
        \includegraphics[width=0.16\textwidth]{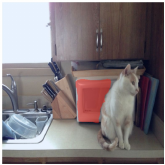} \\
        \includegraphics[width=0.16\textwidth]{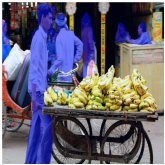} &
        \includegraphics[width=0.16\textwidth]{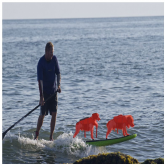} &
        \includegraphics[width=0.16\textwidth]{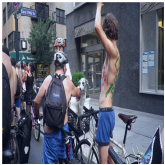} &
        \includegraphics[width=0.16\textwidth]{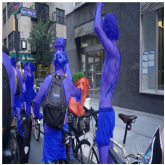} &
        \includegraphics[width=0.16\textwidth]{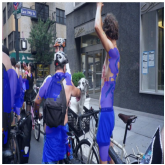} &
        \includegraphics[width=0.16\textwidth]{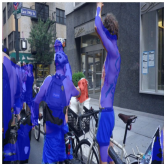} \\
        \midrule
        \includegraphics[width=0.16\textwidth]{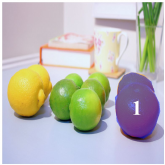} &  
        \includegraphics[width=0.16\textwidth]{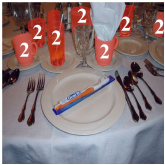} &  
        \includegraphics[width=0.16\textwidth]{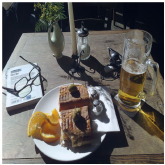} &  
        \includegraphics[width=0.16\textwidth]{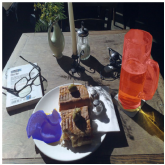} &  
        \includegraphics[width=0.16\textwidth]{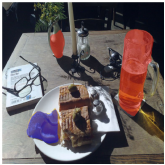} &  
        \includegraphics[width=0.16\textwidth]{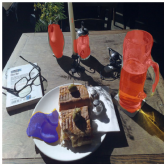} \\
        \includegraphics[width=0.16\textwidth]{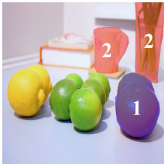} &  
        \includegraphics[width=0.16\textwidth]{figures/qualitative-s2.pdf} &  
        \includegraphics[width=0.16\textwidth]{figures/qualitative-q.pdf} &  
        \includegraphics[width=0.16\textwidth]{figures/qualitative-gt.pdf} &  
        \includegraphics[width=0.16\textwidth]{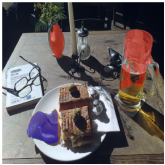} &  
        \includegraphics[width=0.16\textwidth]{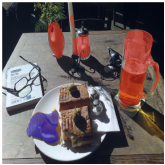} \\
        \includegraphics[width=0.16\textwidth]{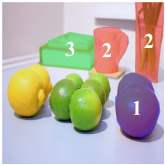} &  
        \includegraphics[width=0.16\textwidth]{figures/qualitative-s2.pdf} &  
        \includegraphics[width=0.16\textwidth]{figures/qualitative-q.pdf} &  
        \includegraphics[width=0.16\textwidth]{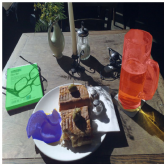} &  
        \includegraphics[width=0.16\textwidth]{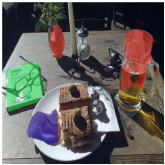} &  
        \includegraphics[width=0.16\textwidth]{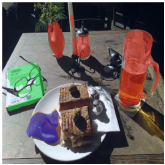}
    \end{tabularx}
    \vspace{-1.4em}
    \caption{\textbf{Qualitative comparison} of CST\textsuperscript{*} \vs our \ours{} on COCO-$20^i$ \cite{Nguyen2019} using 2-way 1-shot tasks. 
    \textbf{(Top)} \emph{Row 1:} Query w/o support classes, \emph{Row 2:} Query w/ subset of support classes, \emph{Row 3:} Query w/ all support classes.
    \textbf{(Bottom)} \emph{Row 1:} Original setting, \emph{Row 2:} Partially augmented setting, \emph{Row 3:} Fully augmented setting.}
    \label{fig:qualitative-results}
    \vspace{-0.6em}
\end{figure}

As explained in \cref{sec:problem-definition}, when handling \textit{N}-way \textit{K}-shot tasks with $N > 1$, the query image can contain \emph{(1)} none, \emph{(2)} some, or \emph{(3)} all of the support classes. 
The top part of \cref{fig:qualitative-results} shows that \ours{} produces more accurate segmentation masks than CST\textsuperscript{*} in these three scenarios, confirming that \ours{} can predict empty masks and masks with one or multiple classes. 
The bottom part of \cref{fig:qualitative-results} illustrates the same task across all few-shot evaluation settings. 
In the original setting, both models segment the \nth{1} class (orange) correctly but make errors on the \nth{2} class (glass), mistakenly segmenting visually similar objects (salt shaker). 
In the partially augmented setting, the additional annotations for the \nth{2} class degrade CST, while \ours{} maintains a precise segmentation, though both still incorrectly segment non-target objects. 
In the fully augmented setting, both methods correctly segment the additional class (book).

\begin{figure}[!t]
    \centering
    \begin{tabularx}{\textwidth}{@{}CCCCCCCCC@{}}
        {\makecell{Full Res. \\ $700\times700$}} &
        {\makecell{Max Res. \\ $50\times50$}} &
        {\makecell{CST$_{l=1}$ \\ $12\times12$}} &
        {\makecell{CST$_{l=2}$ \\ $3\times3$}} &
        {\makecell{\ours$_{l=1}$ \\ $20\times20$}} &
        {\makecell{\ours$_{l=2}$ \\ $10\times10$}} \\
        \includegraphics[width=0.157\textwidth]{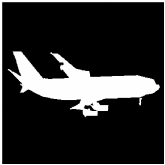} & 
        \includegraphics[width=0.157\textwidth]{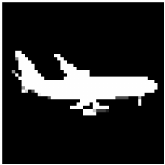} & 
        \includegraphics[width=0.157\textwidth]{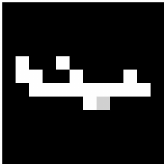} & 
        \includegraphics[width=0.157\textwidth]{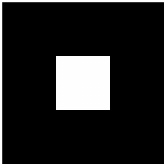} & 
        \includegraphics[width=0.157\textwidth]{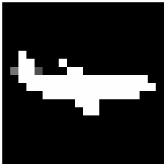} & 
        \includegraphics[width=0.157\textwidth]{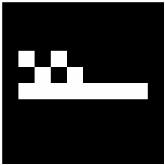} \\
        \includegraphics[width=0.157\textwidth]{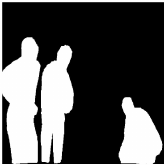} & 
        \includegraphics[width=0.157\textwidth]{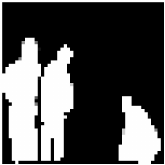} & 
        \includegraphics[width=0.157\textwidth]{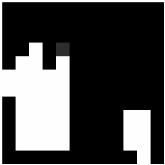} & 
        \includegraphics[width=0.157\textwidth]{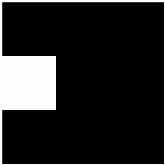} & 
        \includegraphics[width=0.157\textwidth]{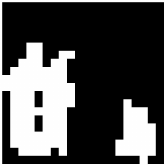} & 
        \includegraphics[width=0.157\textwidth]{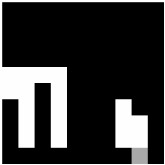}
    \end{tabularx}
    \vspace{-1.4em}
    \caption{\textbf{Segmentation masks used by CST \cite{Kang2023} and \ours{}.} 
    ``Full Res.'' shows the mask at full resolution, while ``Max Res.'' is to the highest resolution compatible with the masked-attention layers of both methods. 
    ``CST$_{l=(\cdot)}$'' and ``\ours$_{l=(\cdot)}$'' indicate the mask resolution used in layer $l \in \{1,2\}$ of CST and \ours, respectively.}
    \label{fig:mask-comparison}
    \vspace{-0.1em}
\end{figure}

To illustrate the effect of higher-resolution tokens, \cref{fig:mask-comparison} compares the segmentation masks used in the masked-attention layers of CST and \ours{}. 
Thanks to our memory-efficient formulation (see \cref{sec:memory-efficient-masked-attention}), \ours{} preserves more details across layers.
The difference is most visible in the second layer, where the mask used by CST barely contains any information since it has a resolution of $3 \times 3$, whereas \ours{}, with a $10 \times 10$ resolution, retains meaningful details and structure. 
For instance, the person in the bottom-right corner of the last row of \cref{fig:mask-comparison} remains visible in both layers of \ours{} but vanishes in the second layer of CST. 
This ability to preserve fine details explains why \ours{} produces more accurate masks especially for small objects, \eg, the dog in the last row of the top part of \cref{fig:qualitative-results}, the handle of the beer glass in the bottom part of \cref{fig:qualitative-results}, and the boat and person in \cref{fig:teaser}.

\subsection{Analysis of Small Objects} \label{sec:analysis-small-objects}
\begin{figure}[!t]
    \centering
    \includegraphics[width=\textwidth]{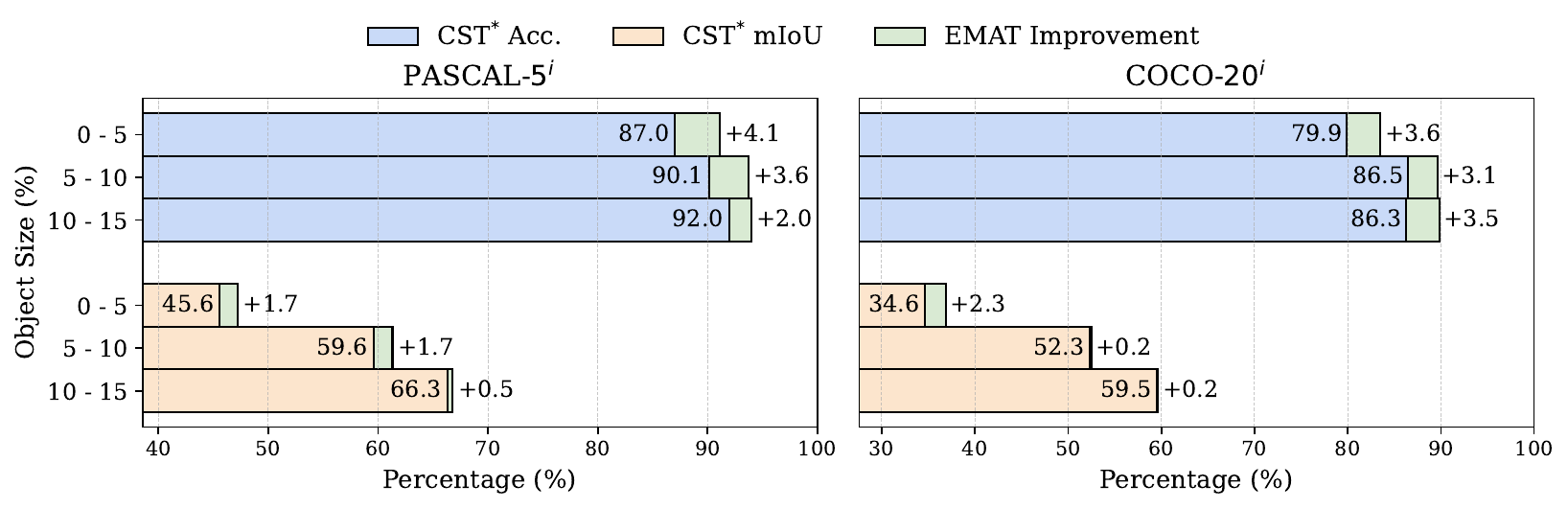}
    \vspace{-2.3em}
    \caption{\textbf{Analysis of small objects} on PASCAL-$5^i$ \cite{Shaban2017} and COCO-$20^i$ \cite{Nguyen2019}. 
    Each bar represents the average across the four folds of each dataset, filtered by object size, using 1-way 1-shot tasks. 
    To enable a more controlled analysis, we modified the original setting (see \cref{sec:problem-definition}) to ensure that the query image always contain the class of the support image. 
    CST\textsuperscript{*} uses the same backbone as \ours{} (\ie, DINOv2 \cite{Oquab2023}).} 
    \label{fig:object-size}
\end{figure}

To further analyze the impact of higher-resolution correlation tokens on small objects, we filter each fold of PASCAL-$5^i$ \cite{Shaban2017} and COCO-$20^i$ \cite{Nguyen2019} based on object size, creating three splits:  objects occupying 0--5\,\%, 5--10\,\%, and 10--15\,\% of the image (see the supplementary material for details on how these splits were defined). 
\cref{fig:object-size} shows the average accuracy and mIoU of CST\textsuperscript{*} and the corresponding improvement achieved by \ours{} across the three splits for both datasets. 
The results indicate that accuracy and mIoU increase with object size, and \ours{} provides the largest improvement over CST\textsuperscript{*} for the smallest objects, gradually decreasing as object size increases. 
The enhanced classification and segmentation accuracy of \ours{} is likely due to better localization enabled by its higher-resolution correlation tokens (see \cref{fig:mask-comparison}).

\begin{table}[!t]
    \centering
    \caption{\textbf{Ablation study of \ours{}} on PASCAL-$5^i$ \cite{Shaban2017} under the original evaluation setting. 
    ``$t^l_s$'' indicates the value of $t_s$ for each layer $l\in\{1,2\}$.
    The memory efficiency (ME), learnable downscaling (LD), and parameter efficiency (PE) columns correspond to the modifications described in \cref{sec:memory-efficient-masked-attention,sec:learnable-downscaling,sec:parameter-efficiency}, respectively.
    ``Mem. Usage'' reports the average per-GPU memory used during training.
    ``All Dataset'' refers to 2-way 1-shot evaluation on the full test set, while ``Small Objects'' restricts evaluation to objects occupying less than 15\,\% of the image, using 1-way 1-shot tasks in which the query always contains the support class. 
    CST\textsuperscript{*} uses the same backbone as \ours{} (\ie, DINOv2 \cite{Oquab2023}). 
    \textbf{(Top)} CST\textsuperscript{*} with its original support dimension per layer $t^l_s$.
    \textbf{(Middle)} CST\textsuperscript{*} with the largest $t^l_s$ that fits in our 48\,GB GPUs.
    \textbf{(Bottom)} successive modifications introduced by \ours{}. 
    \hcolor{} indicates our complete model. 
    \textbf{Bold} and \underline{underlined} values indicate the best and second best results.}
    \label{tab:method-ablation}
    \vspace{-0.5em}
    \begin{tabularx}{\linewidth}{c<{\hspace{0.5em}}l<{\hspace{0.2em}}ccc<{\hspace{0.2em}}CCCCCC}
        \toprule
        \multirow{2}{*}{\makecell{$t^l_s$ \textbf{per} \\ \textbf{Layer}}} &
        \multirow{2}{*}{\textbf{Method}} & 
        \multirow{2}{*}{\textbf{ME}} &
        \multirow{2}{*}{\textbf{LD}} &
        \multirow{2}{*}{\textbf{PE}} &
        \multirow{2}{*}{{\makecell{\textbf{Mem.} \\ \textbf{Usage}}}} &
        \multirow{2}{*}{{\makecell{\textbf{Train.} \\ \textbf{Params.}}}} &
        \multicolumn{2}{c}{\textbf{All Dataset}} &
        \multicolumn{2}{c}{\textbf{Small Objects}} \\
        \cmidrule(lr){8-9}
        \cmidrule(lr){10-11}
        & & & & & & &
        \textbf{Acc.} & 
        {\textbf{mIoU}} &
        \textbf{Acc.} & 
        \textbf{mIoU} \\ 
        \midrule
        \makecell{$t^1_s{=}145$ \\ $t^2_s{=}10$} & CST\textsuperscript{*} & -- & -- & -- & \bfseries 8.68 & \underline{366.00} & 80.58 & 63.28 & 88.96 & 58.16 \\
        \midrule
        \makecell{$t^1_s{=}325$ \\ $t^2_s{=}37$} & CST\textsuperscript{*} & -- & -- & -- & 39.22 & \underline{366.00} & \underline{82.23} & 63.31 & 89.65 & \underline{58.75} \\
        \midrule
        \multirow{4}{*}{\makecell{$t^1_s{=}401$ \\ $t^2_s{=}101$}} & CST\textsuperscript{*} & -- & -- & -- & {\makecell[c]{$\approx $ 63}} & \underline{366.00} & N/A & N/A & N/A & N/A \\
        & \ours & \cmark & -- & -- & 36.92 & \underline{366.00} & 81.95 & 62.97 & 87.99 & 58.06 \\
        & \ours & \cmark & \cmark & -- & \underline{36.53} & 404.48 & 82.17 & \underline{63.36} & \underline{90.49} & 58.73 \\
        & \cellcolor{highlight}\ours{} & \cellcolor{highlight}\cmark & \cellcolor{highlight}\cmark & \cellcolor{highlight}\cmark & \cellcolor{highlight}38.31 & \cellcolor{highlight}\bfseries 86.02 & \cellcolor{highlight}\bfseries 82.70  & \cellcolor{highlight}\bfseries 63.38 & \cellcolor{highlight}\bfseries 91.74 & \cellcolor{highlight}\bfseries 59.17 \\
        \bottomrule
    \end{tabularx}
\end{table}

\subsection{Ablation Study}
\cref{tab:method-ablation} first reports the results of CST\textsuperscript{*} with its original support dimension per layer $t^l_s$. 
For fair comparison, we increased the $t^l_s$ of CST\textsuperscript{*} to use the same as \ours{}, but it required about 63\,GB of GPU memory, which exceeded the 48\,GB capacity of our GPUs, so we instead use the largest $t^l_s$ that fits in our memory. 
For \ours{} we progressively integrated: \emph{(1)} memory-efficient masked attention (see \cref{sec:memory-efficient-masked-attention}), \emph{(2)} learnable downscaling of the query matrix (see \cref{sec:learnable-downscaling}), and \emph{(3)} parameter-efficiency modifications (see \cref{sec:parameter-efficiency}). 
Although \cref{tab:method-ablation} includes results on the full PASCAL-$5^i$ test set, the discussion below focuses on the small-object subset to highlight the effect of each modification introduced by \ours{}.

Adding our memory-efficient masked attention alone lowers memory usage by 26 GB ($\approx$ 41\,\%) but does not improve accuracy or mIoU compared to either variant of CST\textsuperscript{*}, likely because the model relies on large pooling windows for processing the higher-resolution correlation tokens. 
Incorporating our learnable downscaling removes those large windows and yields absolute accuracy gains of +1.53\,\% 
over the original CST\textsuperscript{*} and of +0.84\,\% over the variant with the larger $t^l_s$. It also achieves an absolute mIoU gain of +0.57\,\% compared with the original CST\textsuperscript{*}, while matching the mIoU of the variant with larger $t^l_s$.

Because our learnable downscaling increases the number of trainable parameters, we next apply our parameter-efficiency modifications that remove 318\,K parameters ($\approx$ 79\,\%), while still saving about 39\,\% of the memory CST\textsuperscript{*} would need for using the same $t^l_s$ as \ours{}. 
These modifications result in absolute accuracy gains of +2.78\,\% over the original CST\textsuperscript{*} and +2.09\,\% over the variant with the larger $t^l_s$; mIoU improves by +1.01\,\% and +0.42\,\%, respectively.
\ours{} also slightly improves accuracy and mIoU on the full test set, but its largest gains appear on images containing small objects.

\section{Limitations} 
While the results of \cref{tab:method-ablation} validate that our proposed \ours{} is memory and parameter efficient for high-resolution correlation tokens, its memory efficiency is constrained to datasets where the segmentation masks contain unlabeled areas. 
This limitation arises because our memory-efficient masked attention mechanism is equivalent to self-attention \cite{Vaswani2017,Dosovitskiy2021} when applied to dense semantic-segmentation datasets like Cityscapes \cite{Cordts2016}, where each pixel in the segmentation mask corresponds directly to one of the semantic classes in the image.

\section{Conclusion}
In this work, we propose \ours, an enhancement over CST, the state-of-the-art  method for few-shot classification and segmentation (FS-CS). 
\ours{} incorporates our novel memory-efficient masked attention mechanism that allows our model to process high-resolution correlation tokens while maintaining memory and parameter efficiency. 
Our results demonstrate that \ours{} consistently outperforms all FS-CS methods across all evaluation settings while requiring at least four times fewer trainable parameters. 
Moreover, our qualitative results highlight that \ours{} is capable of correctly generating empty segmentation masks when necessary and capturing finer details more accurately, which improves accuracy when dealing with small objects. 
Additionally, we introduce two novel few-shot evaluation settings designed to maximize the use of the available annotations during inference, reflecting practical few-shot scenarios. 

{\small \inparagraph{Acknowledgments.} 
This work was funded by the Hessian Ministry of Science and Research, Arts and Culture (HMWK) through the
project “The Third Wave of Artificial Intelligence -- 3AI”. 
The work was further supported by the Deutsche Forschungsgemeinschaft (German Research Foundation, DFG) under Germany’s Excellence Strategy (EXC 3057/1 “Reasonable Artificial Intelligence”, Project No.\ 533677015). Stefan Roth acknowledges support by the European Research Council (ERC) under the European Union’s Horizon 2020 research and innovation programme (grant agreement No.\ 866008).}

\clearpage

\appendix
\renewcommand{\thepage}{\roman{page}}
\setcounter{page}{1}

\section{Few-shot Evaluation Settings} 
\begin{figure}[!t]
    \centering
    \includegraphics[width=\textwidth]{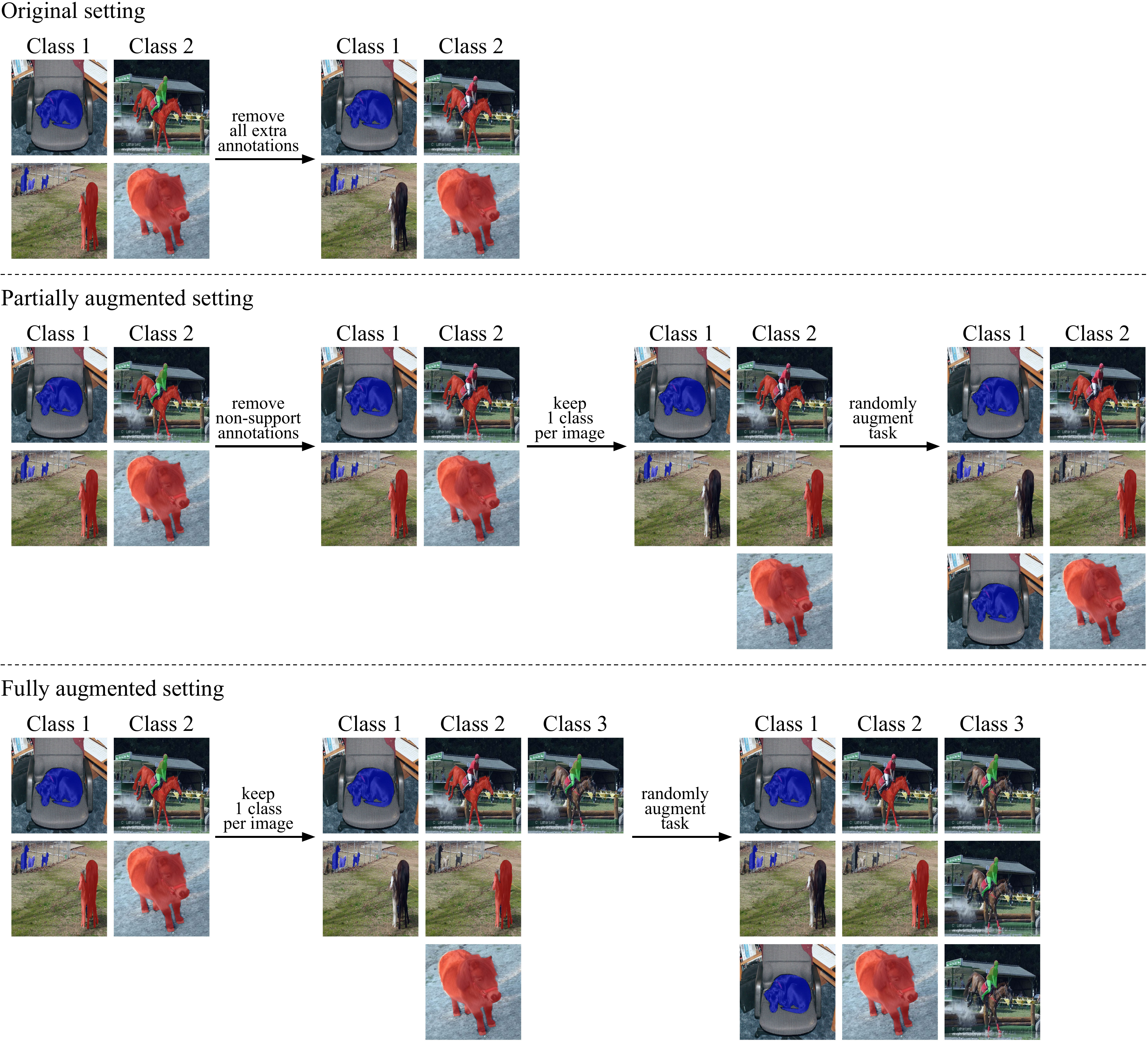}
    \vspace{-2em}
    \caption{\textbf{Task augmentation pipeline for each few-shot evaluation setting.} 
    The 2-way 2-shot (base configuration) task is extracted from the PASCAL-$5^i$ dataset~\cite{Shaban2017}. 
    The \emph{original setting} leaves this configuration unchanged, discarding available annotations. 
    The \emph{partially augmented setting} incorporates unused support-class annotations, yielding a 2-way 3-shot task. 
    The \emph{fully augmented setting} includes all annotations, increasing complexity to 3-way 3-shot.} 
    \label{fig:task-generation}
    \vspace{-0.2em}
\end{figure}

As explained in \cref{sec:proposed-settings}, we introduce two novel few-shot evaluation settings, partially and fully augmented, to better utilize available annotations and represent realistic scenarios. 
\cref{fig:task-generation} illustrates how the support set of a 2-way 2-shot (base task configuration) task is modified under each setting. 

In the \emph{original setting} \cite{Kang2022}, additional annotations are discarded, leaving the base task configuration unchanged. 
The drawback of this setting is that it ignores available annotations. 
For example, in \cref{fig:task-generation}, the second image for the \nth{1} class (dog) also contains annotations for the \nth{2} class (horse), but these are removed.

To address the loss of available annotations, our \emph{partially augmented setting} retains annotations for the support classes (\eg, dog and horse in \cref{fig:task-generation}) while removing those for non-support classes (\eg, person in \cref{fig:task-generation}). 
Since few-shot classification and segmentation (FS-CS) methods treat \textit{N}-way \textit{K}-shot tasks as \textit{N} separate 1-way \textit{K}-shot tasks, each image (shot) must contain only one class (way). 
Therefore, if an image contains multiple support classes, we duplicate it so each copy includes only one. 
This leads to a task configuration that no longer follows the \textit{N}-way \textit{K}-shot definition \cite{Vinyals2016}, as the number of shots per class can vary. 
Consequently, the classes with fewer shots are randomly augmented using the available support data. 
For instance, in \cref{fig:task-generation}, the partially augmented setting converts the 2-way 2-shot into a 2-way 3-shot task. 
It is important to note that in this setting, the number of ways remains fixed but the number of shots can increase up to $N\times K$.

On the other hand, our \emph{fully augmented setting} incorporates all available annotations, including those for non-support classes (\eg, person in \cref{fig:task-generation}). 
The task is then processed similarly to the partially augmented setting. 
However, in this case, both the number of ways and the number of shots can increase, with shots still bounded by $N \times K$. Regardless of the setting, models are evaluated only on the final support classes. 
For example, in \cref{fig:task-generation}, models in the original and partially augmented settings are evaluated on dogs and horses, while in the fully augmented setting, they are also evaluated on people.

\subsubsection{Practical Considerations.} Because annotating data is time- and resource-consuming, few-shot evaluation settings should aim to maximize the use of existing annotations. 
Our partially augmented setting achieves this without increasing task difficulty and may even enhance FS-CS performance by providing more examples per class. 
On the other hand, our fully augmented setting incorporates all available annotations, increasing both the number of ways and the number of shots, making the task more challenging by requiring the model to classify and segment additional classes. 
Nevertheless, neither setting requires extra annotations beyond those already available; instead, they use existing labels discarded by the original setting to better represent realistic scenarios.

Although the loss of annotations in the original setting may not significantly affect training, our partially augmented setting can safely be used in that phase. 
In contrast, our fully augmented setting should be used only for evaluation to prevent data leakage, as it mixes training and test classes within the support set. 
For this reason, we recommend using both settings exclusively for evaluation.  

\subsubsection{Object Size Filtering.} Very small objects can add noise during task augmentation in the partially and fully augmented settings, so we filter out any object that occupy less than a threshold $\theta$ of the image area. 
We set $\theta = 7\,\%$ for PASCAL-$5^i$ \cite{Shaban2017} and $\theta = 3\,\%$ COCO-$20^i$ \cite{Nguyen2019}, values that correspond to the \nth{25} percentile of object sizes in the training set of each dataset. 
These different thresholds reflect dataset-specific object size distribution.

\begin{figure}[!t]
    \centering
    \includegraphics[width=\linewidth]{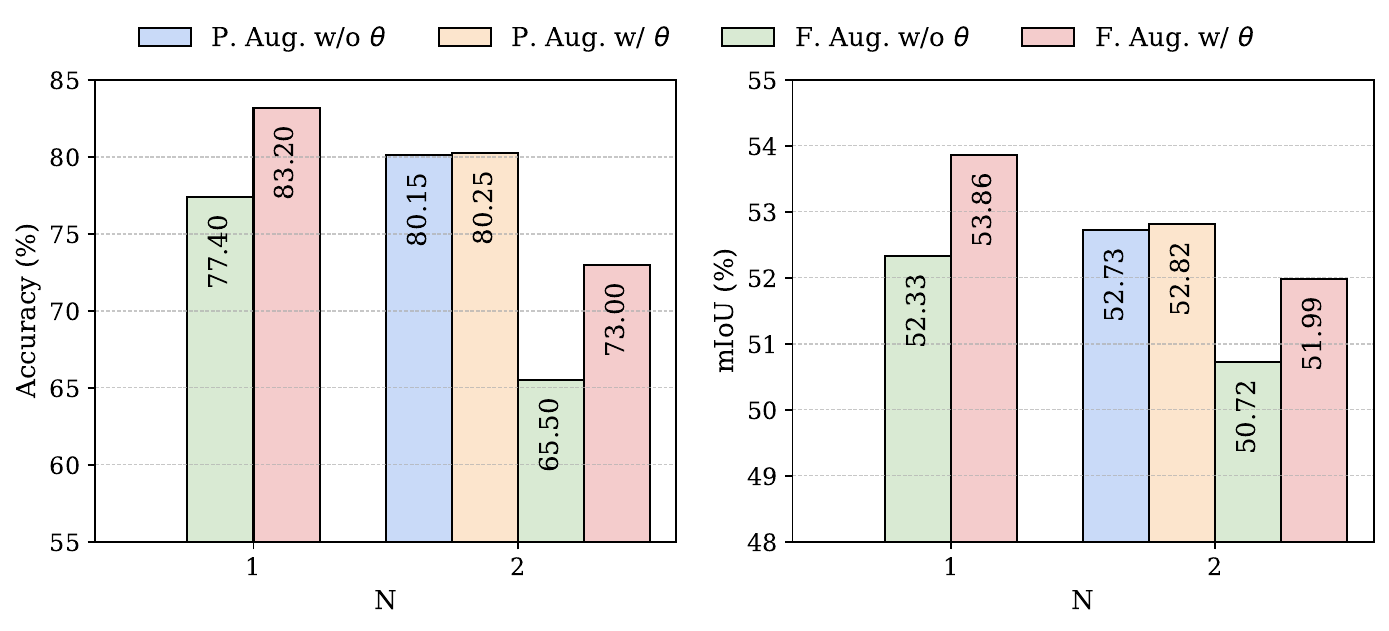}
    \vspace{-2.3em}
    \caption{\textbf{Impact of object size filtering on task augmentation.} 
    ``P.\ Aug.'' and ``F.\ Aug.'' denote the partially  and fully augmented settings, respectively, while $\theta$ indicates whether object size filtering was applied. 
    The bars show the accuracy or mIoU of \ours{} on the COCO-$20^i$ dataset \cite{Nguyen2019} for $\{1,2\}$-way 1-shot tasks. 
    Results for the 1-way 1-shot configuration in the partially augmented setting are omitted becase, when $N = 1$, this setting is equivalent to the original one. In both augmented settings, removing small objects increases accuracy and mIoU.} 
    \label{fig:object-size-filtering}
\end{figure}

\cref{fig:object-size-filtering} shows the impact of applying these thresholds on the classification accuracy and segmentation mIoU of our efficient masked attention transformer (\ours{}) on COCO-$20^i$ for $\{1,2\}$-way 1-shot tasks. 
Filtering out small objects consistently improves accuracy and mIoU in both augmented settings, with a larger gain in the fully augmented setting because it augments more tasks. 
Therefore, all results were obtained using the proposed thresholding.

\section{FS-CS Per-Fold Results} 

\begin{table}[!t]
    \centering
    \caption{\textbf{Per-fold comparison of FS-CS methods on PASCAL-$5^i$ \cite{Shaban2017}.} 
    Results are reported for all evaluation settings (original, partially augmented, and fully augmented) using 1000 tasks per fold with a base configuration of 2-way 1-shot. 
    The number in parentheses in the partially and fully augmented settings indicates the total number of augmented tasks. 
    CST\textsuperscript{*} and \ours{} were trained and evaluated, while other methods were only evaluated using the checkpoints from \cite{Kang2022}. 
    CST\textsuperscript{*} uses the same backbone as \ours{} (\ie, DINOv2 \cite{Oquab2023}). 
    All values are percentages (higher is better). 
    \hcolor{} indicates our proposed method. 
    \textbf{Bold} and \underline{underlined} values represent the best and second best results.}
    \label{tab:per-fold-pascal}
    \vspace{-0.5em}
    \begin{tabularx}{\textwidth}{@{}Xcccccccccc}
        \toprule
        \multirow{2}{*}{\vspace{-0.55em}\textbf{Method}} &
        \multicolumn{2}{c}{\textbf{Fold-0}} &
        \multicolumn{2}{c}{\textbf{Fold-1}} &
        \multicolumn{2}{c}{\textbf{Fold-2}} &
        \multicolumn{2}{c}{\textbf{Fold-3}} &
        \multicolumn{2}{c@{}}{\textbf{Average}} \\ 
        \cmidrule(lr){2-3}
        \cmidrule(lr){4-5}
        \cmidrule(lr){6-7}
        \cmidrule(lr){8-9}
        \cmidrule(lr){10-11} 
        &
        \textbf{Acc.} & \textbf{mIoU} &
        \textbf{Acc.} & \textbf{mIoU} &
        \textbf{Acc.} & \textbf{mIoU} &
        \textbf{Acc.} & \textbf{mIoU} &
        \textbf{Acc.} & \textbf{mIoU} \\ 
        \midrule
        \multicolumn{11}{c}{\textbf{Original}} \\
        \midrule
        PANet \cite{Wang2019}       & 62.30 & 33.31 & 53.30 & 45.94 & 49.50 & 31.20 & 61.00 & 38.34 & 56.53 & 37.20 \\
        PFENet \cite{Tian2020c}      & 23.10 & 31.48 & 53.80 & 46.60 & 40.60 & 31.15 & 39.90 & 33.03 & 39.35 & 35.57 \\
        HSNet \cite{Min2021}        & 68.20 & 43.97 & 73.00 & 55.12 & 56.90 & 35.19 & 71.00 & 45.14 & 67.27 & 44.85 \\
        ASNet \cite{Kang2022}       & 68.20 & 48.44 & 76.20 & 58.19 & 58.80 & 36.45 & 70.00 & 48.41 & 68.30 & 47.87 \\
        CST \cite{Kang2023} & 70.10 & 53.90 & 75.20 & 59.98 & 61.70 & 46.30 & 74.50 & 54.93 & 70.37 & 53.78 \\
        CST\textsuperscript{*} & \underline{89.40} & \underline{64.61} & \underline{80.90} & \textbf{68.16} & \underline{71.40} & \textbf{55.50} & \underline{80.60} & \underline{64.84} & \underline{80.58} & \underline{63.28} \\
        \rowcolor{highlight}
        \ours             & \textbf{90.70} & \textbf{66.68} & \textbf{83.50} & \underline{67.61} & \textbf{72.00} & \underline{54.15} & \textbf{84.60} & \textbf{65.07} & \textbf{82.70} & \textbf{63.38} \\
        \midrule
        \multicolumn{11}{c}{\textbf{Partially Augmented ($\uparrow 242$ tasks)}} \\
        \midrule
        PANet \cite{Wang2019}       & 62.30 & 33.31 & 53.00 & 46.00 & 51.70 & 32.35 & 60.70 & 38.32 & 56.93 & 37.49 \\
        PFENet \cite{Tian2020c}      & 23.10 & 31.48 & 54.10 & 46.68 & 40.70 & 31.28 & 40.00 & 33.02 & 39.48 & 35.61 \\
        HSNet \cite{Min2021}        & 68.20 & 43.97 & 72.90 & 54.85 & 59.00 & 34.90 & 70.90 & 45.18 & 67.75 & 44.72 \\
        ASNet \cite{Kang2022}       & 68.20 & 48.44 & 76.00 & 57.88 & 60.50 & 36.28 & 69.80 & 48.51 & 68.62 & 47.78 \\
        CST \cite{Kang2023} & 70.10 & 53.90 & 75.30 & 60.04 & 62.40 & 46.32 & 74.60 & 54.97 & 70.60 & 53.81 \\
        CST\textsuperscript{*} & \underline{89.40} & \underline{64.61} & \underline{80.90} & \textbf{68.10} & \underline{71.50} & \textbf{55.44} & \underline{80.60} & \underline{64.77} & \underline{80.60} & \underline{63.23} \\
        \rowcolor{highlight}
        \ours                & \textbf{90.70} & \textbf{66.68} & \textbf{83.30} & \underline{67.48} & \textbf{73.00} & \underline{54.03} & \textbf{84.70} & \textbf{65.10} & \textbf{82.92} & \textbf{63.32} \\
        \midrule
        \multicolumn{11}{c}{\textbf{Fully Augmented ($\uparrow 694$ tasks)}} \\
        \midrule
        PANet \cite{Wang2019}       & 62.30 & 33.31 & 51.70 & 45.95 & 49.60 & 31.65 & 59.40 & 38.11 & 55.75 & 37.25 \\
        PFENet \cite{Tian2020c}      & 23.10 & 31.48 & 52.60 & 46.14 & 33.70 & 30.33 & 38.10 & 32.35 & 36.88 & 35.08 \\
        HSNet \cite{Min2021}        & 68.20 & 43.97 & 72.20 & 54.55 & 53.80 & 33.87 & 69.50 & 45.19 & 65.92 & 44.40 \\
        ASNet \cite{Kang2022}       & 68.20 & 48.44 & 75.00 & 57.66 & 54.30 & 35.90 & 68.10 & 48.31 & 66.40 & 47.58 \\
        CST \cite{Kang2023} & 70.10 & 53.90 & 74.50 & 59.50 & 56.70 & 46.62 & 72.50 & 55.01 & 68.45 & 53.76 \\
        CST\textsuperscript{*} & \underline{89.40} & \underline{64.61} & \underline{80.20} & \textbf{67.43} & \underline{66.40} & \textbf{55.80} & \underline{78.30} & \underline{64.47} & \underline{78.57} & \underline{63.08} \\
        \rowcolor{highlight}
        \ours                & \textbf{90.70} & \textbf{66.68} & \textbf{82.60} & \underline{66.93} & \textbf{68.30} & \underline{54.37} & \textbf{83.30} & \textbf{65.00} & \textbf{81.23} & \textbf{63.24} \\
        \bottomrule
    \end{tabularx}
    \vspace{-1.3em}
\end{table}

\cref{sec:comparison-sota-fscs} compares our \ours{} with current state-of-the-art (SOTA) FS-CS methods, reporting results averaged over the four folds of the PASCAL-$5^i$ and COCO-$20^i$ datasets using 2-way 1-shot tasks. 
To complement those results, \cref{tab:per-fold-pascal,tab:per-fold-coco} present per-fold results, consistently showing that \ours{} outperforms all FS-CS methods across most folds, datasets, and evaluation settings. 
These tables also report the number of augmented tasks in our proposed evaluation settings. 
On PASCAL-$5^i$, the partially and fully augmented settings augment 242 and 694 tasks, respectively, meaning that about 6\,\% of tasks contain extra support-class annotations (partially augmented), and 11\,\% contain extra annotations for non-support classes (fully augmented). 
On COCO-$20^i$, the corresponding augmentation rates are roughly 3\,\% and 35\,\%. 
These results confirm that our evaluation settings use annotations discarded by the original setting, thereby creating more realistic evaluation scenarios.

\begin{table}[!t]
    \centering
    \caption{\textbf{Per-fold comparison of FS-CS methods on COCO-$20^i$ \cite{Nguyen2019}.} 
    Results are reported for all evaluation settings (original, partially augmented, and fully augmented) using 1000 tasks per fold with a base configuration of 2-way 1-shot. 
    The number in parentheses in the partially and fully augmented settings indicates the total number of augmented tasks. 
    CST\textsuperscript{*} and \ours{} were trained and evaluated, while other methods were only evaluated using the checkpoints from \cite{Kang2022}. 
    CST\textsuperscript{*} uses the same backbone as \ours{} (\ie, DINOv2 \cite{Oquab2023}). 
    All values are percentages (higher is better). 
    \hcolor{} indicates our proposed method. 
    \textbf{Bold} and \underline{underlined} values represent the best and second best results.}
    \label{tab:per-fold-coco}
    \vspace{-0.5em}
    \begin{tabularx}{\textwidth}{@{}Xcccccccccc}
        \toprule
        \multirow{2}{*}{\vspace{-0.55em}\textbf{Method}} &
        \multicolumn{2}{c}{\textbf{Fold-0}} &
        \multicolumn{2}{c}{\textbf{Fold-1}} &
        \multicolumn{2}{c}{\textbf{Fold-2}} &
        \multicolumn{2}{c}{\textbf{Fold-3}} &
        \multicolumn{2}{c@{}}{\textbf{Average}} \\ 
        \cmidrule(lr){2-3}
        \cmidrule(lr){4-5}
        \cmidrule(lr){6-7}
        \cmidrule(lr){8-9}
        \cmidrule(lr){10-11} 
        &
        \textbf{Acc.} & \textbf{mIoU} &
        \textbf{Acc.} & \textbf{mIoU} &
        \textbf{Acc.} & \textbf{mIoU} &
        \textbf{Acc.} & \textbf{mIoU} &
        \textbf{Acc.} & \textbf{mIoU} \\ 
        \midrule
        \multicolumn{11}{c}{\textbf{Original}} \\
        \midrule
        PANet \cite{Wang2019}       & 46.60 & 24.91 & 52.70 & 24.98 & 55.90 & 23.31 & 50.00 & 21.36 & 51.30 & 23.64 \\
        PFENet \cite{Tian2020c}      & 35.60 & 23.99 & 34.30 & 24.57 & 43.10 & 20.99 & 32.80 & 23.93 & 36.45 & 23.37 \\
        HSNet \cite{Min2021}        & 57.70 & 29.77 & 62.30 & 30.94 & 67.10 & 31.31 & 62.60 & 30.31 & 62.43 & 30.58 \\
        ASNet \cite{Kang2022}       & \underline{59.50} & 29.75 & 61.50 & 32.99 & 68.80 & 33.41 & 62.40 & 30.35 & 63.05 & 31.62 \\
        CST \cite{Kang2023} & 61.00 & 34.74 & 66.40 & 37.14 & 68.20 & 36.76 & 60.50 & 36.29 & 64.02 & 36.23 \\
        CST\textsuperscript{*} & \textbf{74.00} & \underline{49.38} & \underline{79.90} & \underline{53.88} & \underline{81.30} & \underline{51.46} & \underline{79.60} & \underline{51.15} & \underline{78.70} & \underline{51.47} \\
        \rowcolor{highlight}
        \ours             & \textbf{74.00} & \textbf{50.54} & \textbf{83.10} & \textbf{55.44} & \textbf{83.10} & \textbf{53.05} & \textbf{80.10} & \textbf{52.19} & \textbf{80.07} & \textbf{52.81} \\
        \midrule
        \multicolumn{11}{c}{\textbf{Partially Augmented ($\uparrow 106$ tasks)}} \\
        \midrule
        PANet \cite{Wang2019}       & 46.40 & 25.27 & 52.80 & 25.10 & 55.90 & 23.27 & 50.20 & 21.48 & 51.32 & 23.78 \\
        PFENet \cite{Tian2020c}      & 35.80 & 24.08 & 34.30 & 24.52 & 43.00 & 21.00 & 32.90 & 23.96 & 36.50 & 23.39 \\
        HSNet \cite{Min2021}        & 57.50 & 29.88 & 62.60 & 31.02 & 67.00 & 31.30 & 62.50 & 30.44 & 62.40 & 30.66 \\
        ASNet \cite{Kang2022}       & 59.30 & 29.87 & 61.70 & 32.95 & 68.80 & 33.39 & 62.30 & 30.36 & 63.03 & 31.64 \\
        CST \cite{Kang2023} & 61.30 & 34.59 & 66.20 & 37.08 & 68.40 & 36.76 & 60.50 & 36.36 & 64.10 & 36.20 \\
        CST\textsuperscript{*} & \textbf{74.60} & \underline{49.27} & \underline{79.80} & \underline{53.98} & \underline{81.50} & \underline{51.68} & \underline{79.60} & \underline{51.19} & \underline{78.87} & \underline{51.53} \\
        \rowcolor{highlight}
        \ours               & \underline{74.30} & \textbf{50.59} & \textbf{83.20} & \textbf{55.42} & \textbf{83.30} & \textbf{53.06} & \textbf{80.20} & \textbf{52.22} & \textbf{80.25} & \textbf{52.82} \\
        \midrule
        \multicolumn{11}{c}{\textbf{Fully Augmented ($\uparrow 1515$ tasks)}} \\
        \midrule
        PANet \cite{Wang2019}       & 30.90 & 24.43 & 49.30 & 24.44 & 53.50 & 22.98 & 46.60 & 20.81 & 45.07 & 23.17 \\
        PFENet \cite{Tian2020c}      & 19.60 & 20.33 & 29.10 & 23.79 & 40.70 & 20.40 & 27.90 & 21.92 & 29.33 & 21.61 \\
        HSNet \cite{Min2021}        & 40.20 & 27.73 & 57.50 & 29.82 & 64.30 & 30.96 & \underline{58.60} & 29.26 & 55.15 & 29.44 \\
        ASNet \cite{Kang2022}       & 41.40 & 27.23 & 55.70 & 32.10 & 66.60 & 33.14 & 58.20 & 29.43 & 55.47 & 30.47 \\
        CST \cite{Kang2023} & 44.70 & 33.74 & 59.70 & 36.32 & 65.60 & 36.87 & 55.20 & 35.46 & 56.30 & 35.60 \\
        CST\textsuperscript{*} & \underline{54.40} & \underline{48.25} & \underline{74.80} & \underline{52.86} & \underline{79.70} & \underline{51.69} & \textbf{75.80} & \underline{50.25} & \underline{71.18} & \underline{50.76} \\
        \rowcolor{highlight}
        \ours               & \textbf{56.30} & \textbf{49.53} & \textbf{78.50} & \textbf{54.46} & \textbf{81.40} & \textbf{53.06} & \textbf{75.80} & \textbf{50.90} & \textbf{73.00} & \textbf{51.99} \\
        \bottomrule
    \end{tabularx}
\end{table}

\begin{figure}[!t]
    \centering
    \includegraphics[width=\linewidth]{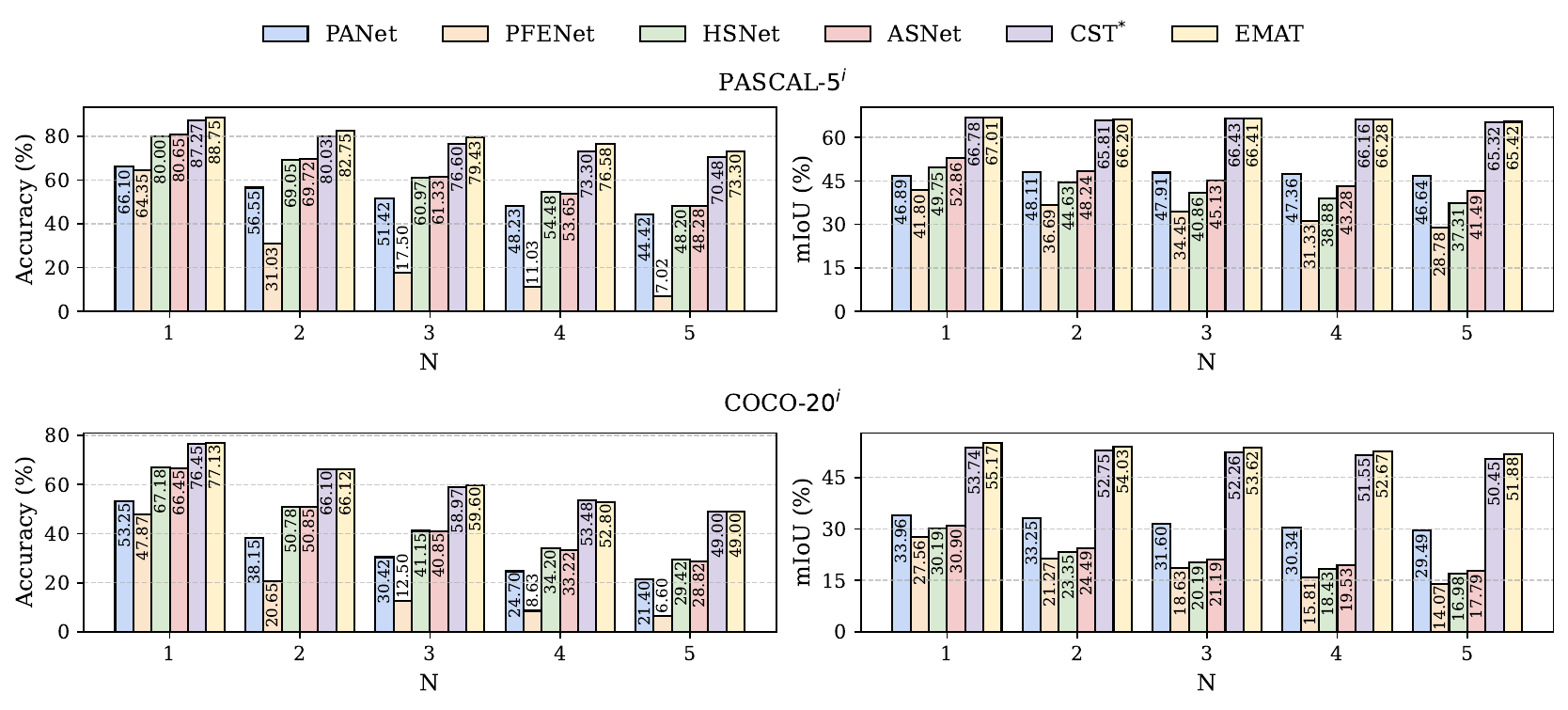}
    \vspace{-2em}
    \caption{\textbf{Comparison of FS-CS methods} on PASCAL-$5^i$ \cite{Shaban2017} and COCO-$20^i$ \cite{Nguyen2019} under the fully augmented setting. 
    CST\textsuperscript{*} uses the same backbone as \ours{} (\ie, DINOv2 \cite{Oquab2023}). 
    Results are averaged over 4000 tasks (1000 per fold) using a base task configuration of $\{1,\dots,5\}$-way 5-shot. 
    The number of augmented tasks for each value of \textit{N} in PASCAL-$5^i$ is 1243, 1758, 2094, 2319, and 2522, respectively. 
    For COCO-$20^i$, the corresponding values are 2136, 2982, 3432, 3674, and 3806.} 
    \label{fig:n-way-5-shot-comparison}
\end{figure}

To evaluate the scalability across different \textit{N}-way \textit{K}-shot configurations, \cref{fig:n-way-5-shot-comparison} shows the accuracy and mIoU of all FS-CS methods on PASCAL\nobreakdash-$5^i$ and COCO-$20^i$ in the fully augmented setting using $\{1,\dots,5\}$-way 5-shot tasks. 
\ours{} consistently achieves higher classification accuracy and mIoU compared to other FS-CS methods. 
Notably, while increasing the number of classes (\textit{N}) typically raises task difficulty, \ours{} maintains stable segmentation performance, further validating the fully augmented setting as a challenging yet realistic and effective benchmark for FS-CS evaluation.

\section{Analysis of Object Size Distribution in PASCAL-$5^i$}

\begin{figure}[!t]
    \centering
    \includegraphics[width=\textwidth]{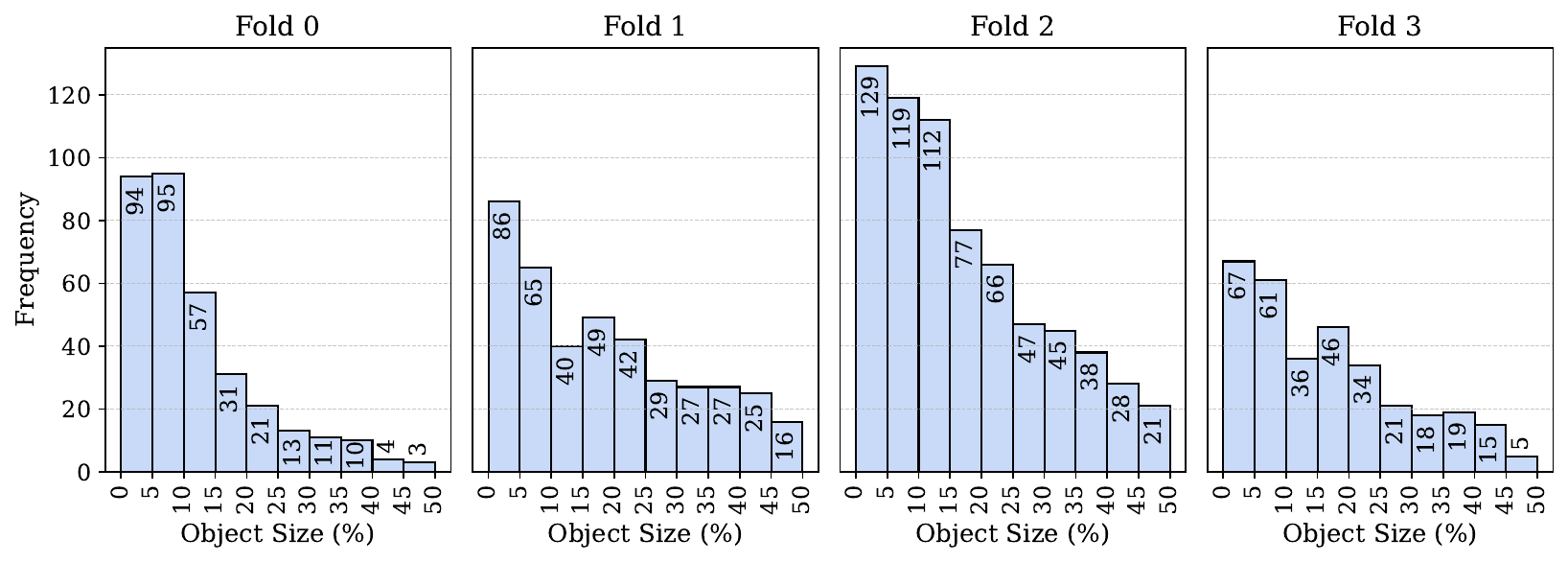} 
    \vspace{-2em}
    \caption{\textbf{Per-fold object size distribution in the PASCAL-$5^i$ \cite{Shaban2017} test set}. Each histogram shows the frequency of objects based on their relative size (as a percentage of the image area), up to 50\,\%.}
    \label{fig:size-distribution}
\end{figure}

\cref{sec:analysis-small-objects} analyzes the impact of higher-resolution correlation tokens on small objects by filtering each fold of PASCAL-$5^i$ and COCO-$20^i$ based on object size. 
This filtering results in three splits: objects occupying 0--5\,\%, 5--10\,\%, and 10--15\,\% of the image. 
These intervals were chosen after examining the object size distribution in the PASCAL-$5^i$ test set, which has fewer test examples than COCO-$20^i$. 

\cref{fig:size-distribution} presents frequency histograms of object sizes (up to 50\,\% of the image) for all four PASCAL-$5^i$ test folds. 
Based on the fold with more examples (Fold 2), we selected the size intervals that ensure each split contains at least 100 examples, \ie, 0--5\,\%, 5--10\,\%, and 10--15\,\%. 
For consistency and comparability, we apply the same splits to COCO-$20^i$. 
Furthermore, \cref{fig:size-distribution} shows that the majority of objects in PASCAL-$5^i$ are very small, highlighting the importance of accurately classifying and segmenting small objects, a challenge that our \ours{} effectively addresses.

\section{Comparison to SOTA FS-S}

\begin{table}[!t]
    \centering
    \caption{\textbf{Comparison of FS-S methods} on PASCAL-$5^i$ \cite{Shaban2017} and COCO-$20^i$ \cite{Nguyen2019} using 1-way $\{1,5\}$-shot tasks. 
    For fairness, \ours$_s$ exclude the classification head. 
    All values correspond to mIoU in percentage (higher is better). 
    The mIoU values for \ours$_s$ were obtained through our own training and evaluation, while values for other methods are taken directly from their respective papers. 
    The top part of the table shows methods based on ResNet backbones, and the bottom part shows those based on foundation models. 
    \hcolor{} indicates our proposed method. 
    \textbf{Bold} and \underline{underlined} values represent the best and second best results.}
    \label{tab:fss-comparison}
    \vspace{-0.5em}
    \begin{tabularx}{\textwidth}{@{}llCcccc}
        \toprule
        \multirow{2}{*}{\vspace{-0.55em}\textbf{Method}} &
        \multirow{2}{*}{\vspace{-0.55em}\textbf{Venue}} &
        \multirow{2}{*}{\vspace{-0.55em}\textbf{Backbone}} &
        \multicolumn{2}{c}{\textbf{PASCAL-$5^i$}} &
        \multicolumn{2}{c}{\textbf{COCO-$20^i$}} \\ 
        \cmidrule(lr){4-5}
        \cmidrule(lr){6-7} 
        & & & 
        \textbf{1-s} & \textbf{5-s} &
        \textbf{1-s} & \textbf{5-s} \\ 
        \midrule
        MIANet \cite{Yang2023} & CVPR'23 & ResNet50 & 68.7 & 71.6 & 47.7 & 51.7 \\
        HDMNet \cite{Peng2023} & CVPR'23 & ResNet50 & 69.4 & 71.8 & 50.0 & 56.0 \\
        VAT + MSI \cite{Moon2023} & ICCV'23 & ResNet101 & 70.1 & 72.2 & 49.8 & 54.0 \\
        SCCAN \cite{Xu2023} & ICCV'23 & ResNet101 & 68.3 & 71.5 & 48.2 & 57.0 \\
        AMFomer \cite{Wang2023} & NeurIPS'23 & ResNet101 & 70.7 & 73.6 & 51.0 & 57.3 \\
        PMNet \cite{Chen2024b} & WACV'24 & ResNet101 & 68.1 & 73.9 & 43.7 & 53.1 \\
        ABCB \cite{Zhu2024b} & CVPR'24 & ResNet101 & \underline{72.0} & \underline{74.9} & \underline{51.5} & \underline{58.8} \\
        \rowcolor{highlight}
        \ours$_s$ & -- & DINOv2-S & \textbf{72.5} & \textbf{75.9} & \textbf{59.8} & \textbf{65.0} \\
        \midrule
        LLaFS \cite{Zhu2024} & CVPR'24 & ResNet50 + CodeLlama-7B & \underline{73.5} & 75.6 & 53.9 & 60.0 \\
        PI-CLIP \cite{Wang2024} & CVPR'24 & ResNet50 + CLIP-B & \textbf{76.8} & \underline{77.2} & 56.8 & 59.1 \\
        Matcher \cite{Liu2024} & ICLR'24 & DINOv2-L + SAH & 68.1 & 74.0 & 52.7 & 60.7 \\
        GF-SAM \cite{Zhang2024} & NeurIPS'24 & DINOv2-L + SAH & 72.1 & \textbf{82.6} & \underline{58.7} & \textbf{66.8} \\
        \rowcolor{highlight}
        \ours$_s$ & -- & DINOv2-S & 72.5 & 75.9 & \textbf{59.8} & \underline{65.0} \\
        \bottomrule
    \end{tabularx}
\end{table}

To compare our \ours{} with SOTA few-shot segmentation (FS-S) methods, we removed its classification head and denote this adapted version as \ours$_s$. 
This adaptation ensures a fair comparison to existing FS-S methods, as simultaneously handling classification and segmentation is more challenging than focusing only on segmentation. 
Furthermore, we adopt the 1-way \textit{K}-shot formulation used by the FS-S methods, where the query image always contains the support class. 

\cref{tab:fss-comparison} presents a comparison between \ours$_s$ and SOTA FS-S methods, divided into two groups: methods based on ResNet backbones and those based on foundation models. 
Although \ours$_s$ is based on a foundation model (DINOv2), it uses the small variant (DINOv2-S), whose model size is comparable to ResNet50. 
Therefore, comparing \ours$_s$ with FS-S methods based on ResNet backbones is still meaningful. 

The top part of \cref{tab:fss-comparison} shows that \ours$_s$ consistently outperforms all ResNet-based FS-S methods across both PASCAL-$5^i$ and COCO-$20^i$ datasets and across all task configurations, even surpassing methods based on a significantly larger backbone, ResNet101. 
Notably, the improvement in mIoU is more pronounced on COCO-$20^i$, with absolute gains of +8.3\,\% and +6.2\,\% for 1-shot and 5-shot settings, respectively.

On the other hand, comparing \ours$_s$ to FS-S methods based on foundation models (bottom part of \cref{tab:fss-comparison}) is less straightforward for two main reasons: \emph{(1)} all baseline methods combine two models, and \emph{(2)} the foundation models they use are significantly larger than the one used by \ours$_s$. 
For example, both Matcher \cite{Liu2024} and GF-SAM \cite{Zhang2024} use the large version of DINOv2 (\ie, DINOv2-L), while \ours$_s$ uses the small variant (\ie, DINOv2-S). 
Nevertheless, our method still achieves an absolute improvement of +1.1\,\% mIoU on COCO-$20^i$ in the 1-way 1-shot setting and overall obtains competitive mIoU compared to other SOTA FS-S methods based on large foundation models.

\end{document}